\documentclass[sn-mathphys,Numbered]{sn-jnl}

\usepackage{graphicx}%
\usepackage{multirow}%
\usepackage{amsmath,amssymb,amsfonts}%
\usepackage{amsthm}%
\usepackage{mathrsfs}%
\usepackage[title]{appendix}%
\usepackage{xcolor}%
\usepackage{textcomp}%
\usepackage{manyfoot}%
\usepackage{booktabs}%
\usepackage{algorithm}%
\usepackage{algorithmicx}%
\usepackage{algpseudocode}%
\usepackage{listings}%
\usepackage{subfig}
\usepackage{adjustbox}

\theoremstyle{thmstyleone}%
\theoremstyle{thmstyletwo}%

\theoremstyle{thmstylethree}%

\begin{document}

\title[Article Title]{A Mountain-Shaped Single-Stage Network for Accurate Image Restoration}


\author[1]{\fnm{Hu} \sur{Gao}}\email{gao\_h@mail.bnu.edu.cn}
\author[1]{\fnm{Jing} \sur{Yang}}\email{yangjing2019@mail.bnu.edu.cn}
\author[1]{\fnm{Ying} \sur{Zhang}}\email{202120081049@mail.bnu.edu.cn}
\author[1]{\fnm{Ning} \sur{Wang}}\email{wangningbnu@mail.bnu.edu.cn}
\author[1]{\fnm{Jingfan} \sur{Yang}}\email{yangjingfan@mail.bnu.edu.cn}
\author*[1]{\fnm{Depeng} \sur{Dang}}\email{ddepeng@bnu.edu.cn}

\affil[1]{\orgdiv{School of Artificial Intelligence}, \orgname{Beijing Normal University}, \orgaddress{\city{Beijing}, \postcode{100000}, \country{China}}}


\abstract{Image restoration is the task of aiming to obtain a high-quality image from a corrupt  input image, such as  deblurring and  deraining. In image restoration, it is typically necessary to maintain a complex balance between spatial details and contextual information. Although a multi-stage network can optimally balance these competing goals and achieve significant performance, this also increases the system's complexity. In this paper, we propose a mountain-shaped single-stage design base on a simple U-Net architecture, which removes or replaces unnecessary nonlinear activation functions to achieve the above balance with low system complexity. Specifically, we propose a feature fusion middleware (FFM) mechanism as an information exchange component between the encoder-decoder architectural levels. It seamlessly integrates upper-layer information into the adjacent lower layer, sequentially down to the lowest layer. Finally, all information is fused into the original image resolution manipulation level. This preserves spatial details and integrates contextual information, ensuring high-quality image restoration. In addition, we propose a multi-head attention middle block (MHAMB) as a bridge between the encoder and decoder to capture more global information and surpass the limitations of the receptive field of CNNs. Extensive experiments demonstrate that our approach, named as M3SNet, outperforms previous state-of-the-art models while using less than half the computational costs, for several image restoration tasks, such as image deraining and deblurring.}




\maketitle

\section{Introduction}\label{sec1}
\begin{figure}[b] 
	\centering
	\includegraphics[width=1\textwidth]{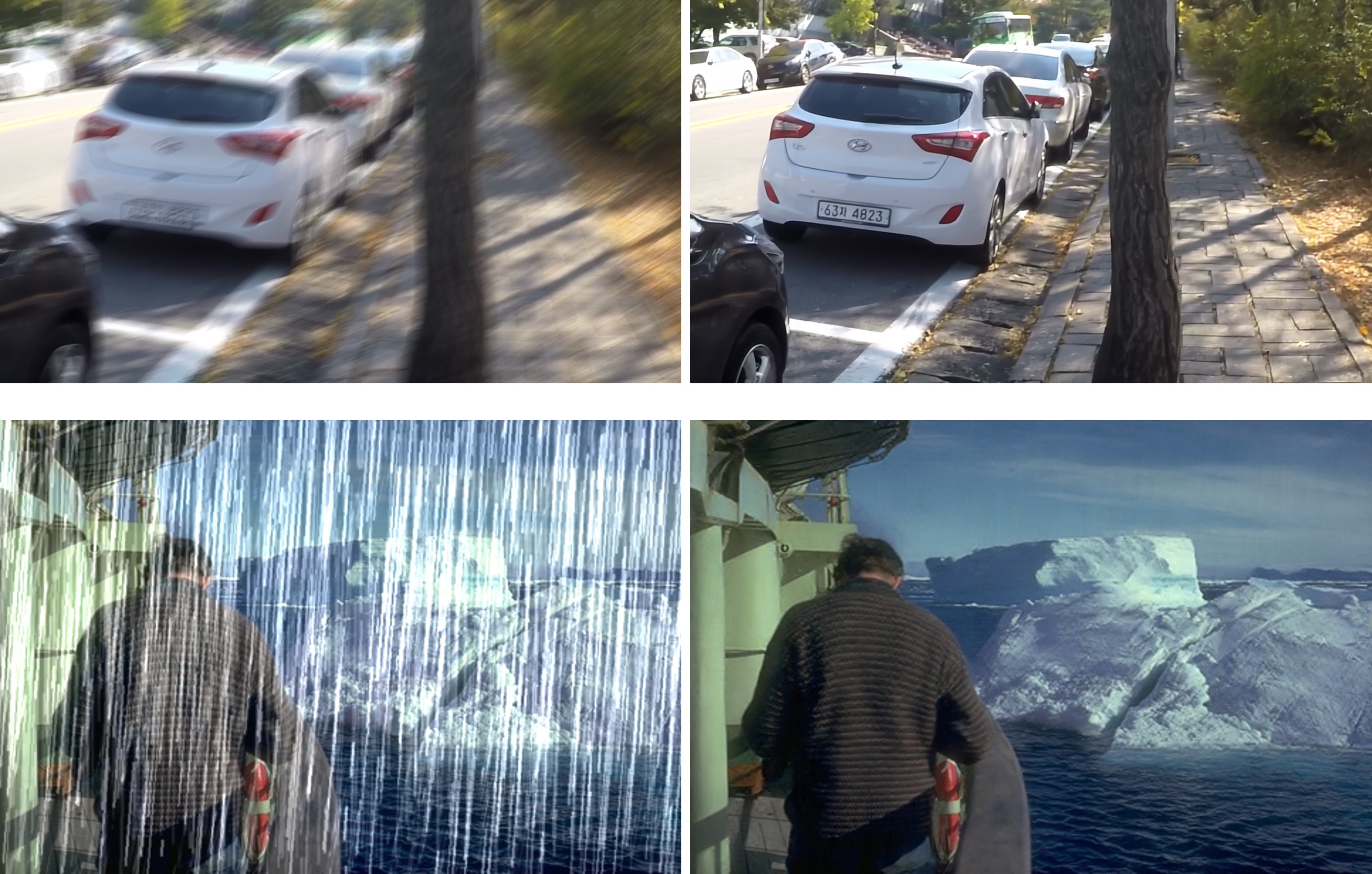}
	\caption{Visualized results of M3SNet on various image
restoration tasks. Left: degraded image. Right: the predicted result of M3SNet. From top to bottom: image deblurring, and image deraining task respectively.}
	\label{fig:001}
\end{figure}
Image degradation is a common issue that occurs during image acquisition due to a variety of factors such as camera limitations, environmental conditions, and human factors. For instance, smartphone cameras with narrow apertures, small sensors, and limited dynamic range can produce blurred and noisy images due to device shaking caused by body movements. Similarly, images captured in adverse weather conditions can be affected by haze and rain.  Most classical image restoration tasks can be formulated as:
\begin{equation}
	\label{equ:001}
	\mathbf{L} = \mathbf{D}(\mathbf{H}) + \gamma
\end{equation}
where $\mathbf{L}$ denotes an observed low-quality image, $\mathbf{H}$ refers to its corresponding high-quality image, and $\mathbf{D}(\cdot), \gamma $indicate the degradation function and the noise during the imaging and transmission processes, respectively. This formulation can signify different image restoration tasks when $\mathbf{D}(\cdot)$ varies. For example, if $\mathbf{D(H)} = \mathbf{H}t - At + A$, it corresponds to the deraining or dehazing task, where $A$ is the global atmospheric light and $t$ is the transmission matrix defined as:
\begin{equation}
	\label{equ:002}
	t = e^{-\alpha d}
\end{equation}
where $\alpha$ is the scattering coefficient of the atmosphere, and $d$ is the distance between the object and the camera.

Image restoration  aims to recover the high-quality clean image $\mathbf{H}$ from its degraded image $\mathbf{L}$. It is a highly ill-posed problem as there are many candidates for any original input. In order to restrict the infinite feasible candidates space to natural images, traditional methods~\cite{1992Nonlinear, 1997Prior, 2002Scale, 2005Fields, 2010Single, 2011Image, 2011Single} explicitly design appropriately priors for the given kind of restoration problem, such as domain-relevant priors and task-relevant priors. Then, the potential high-quality image can be obtained by solving a maximum a posteriori (MAP) problem:
\begin{equation}
	\label{equ:02}
	\mathbf{\hat{H}}= \underset {\mathbf{H}} { \operatorname {arg\,max}} \log P(\mathbf{L}|\mathbf{H}) + \log P(\mathbf{H})
\end{equation}
where $P(\mathbf{L}|\mathbf{H})$ represents the probability of observing the degraded image $\mathbf{L}$ given the clean image $\mathbf{H}$, and $P(\mathbf{H})$ represents the prior distribution of the clean image $\mathbf{H}$. This can also be expressed as a constrained maximum likelihood estimation:
\begin{equation}
	\label{equ:021}
	\mathbf{\hat{H}}= \underset {\mathbf{H}} { \operatorname {arg\,min}} \ \| \mathbf{L}-\mathbf{D(H)} \|^2 + \lambda\Psi(\mathbf{H})
\end{equation}
where fidelity term $| \mathbf{L}-\mathbf{D(H)} |^2$ serves as an approximation for the likelihood $P(\mathbf{L}|\mathbf{H})$, while the regularization term $\lambda\Psi(\mathbf{H})$ represents either the priors of the latent image $\mathbf{H}$ or the constraints on the solution. The aim is to express the fidelity of the reconstructed image to the original input while simultaneously considering the prior knowledge or constraints imposed on the solution

While designing effective priors for image restoration can be challenging and may not be generalizable.  With large-scale data, deep models such as Convolutional Neural Networks(CNNs)~\cite{Zamir2021MPRNet, Chen_2021_CVPR,PREnet,RESCAN,chen2022simple,chu2022nafssr, 2022Learning} and Transformer~\cite{zhang2023accurate, Zamir2021Restormer, Tsai2022Stripformer,Wang_2022_CVPR} have emerged as the preferred choice due to their ability to implicitly learn more general priors by capturing natural image statistics and achieving state-of-the-art (SOTA) performance in image restoration. The performance gain of these deep learning models over conventional restoration approaches is primarily attributed to their model design, which includes numerous network modules and functional units for image restoration, such as recursive residual learning~\cite{Wang2018ESRGANES}, transformer~\cite{Zamir2021Restormer, zhang2023accurate, Wang_2022_CVPR}, encoder-decoders~\cite{RFB15a,chen2022simple,chu2022nafssr}, multi-scale models~\cite{Mssnet,9156921}, and generative models~\cite{Degan,DBGAN,deganv2}.

Nevertheless, most of these models for low-level vision problems are based on a single-stage design, which ignores the interactions that exist between spatial details and contextualized information. To address this limitation,~\cite{Zamir2021MPRNet, Chen_2021_CVPR,PREnet,RESCAN} proposes a multi-stage architecture in which contextualized features are first learned through an encoder-decoder architecture and subsequently integrated with a high-resolution branch to preserve local information. Despite its good performance, this method requires refining the results from the previous stage in the later stage, leading to a high level of system complexity.

Based on the information presented, a natural question that comes to mind is whether it is feasible to use a single-stage architecture to reduce system complexity and  achieve the same balance between spatial details and contextualized information as a multi-stage architecture while  maintaining the SOTA performance. To achieve this objective, we propose a mountain-shaped single-stage image restoration architecture, called M3SNet, with several key components. 1). We utilize NAFNet~\cite{chen2022simple} as the baseline architecture and concentrate on modifying the network model to attain multi-stage functionality. By emitting the information transfer between the multi-stage and eliminating the nonlinear activation function from the network structure, we are able to reduce the system's complexity. 2). A feature fusion middleware (FFM) mechanism has been added to facilitate multi-scale information fusion between encoder and decoder blocks from different layers, resulting in the acquisition of more contextual information. Additionally, this approach enables manipulation of the original image resolution, thereby aiding in the preservation of spatial details.
3). A multi-head attention middle block (MHAMB) is the bridge between the encoder and decoder that  surpass the limitations of the receptive field of CNNs and capture more global information.

\begin{figure}[t] 
	\centering
	\includegraphics[width=1\textwidth]{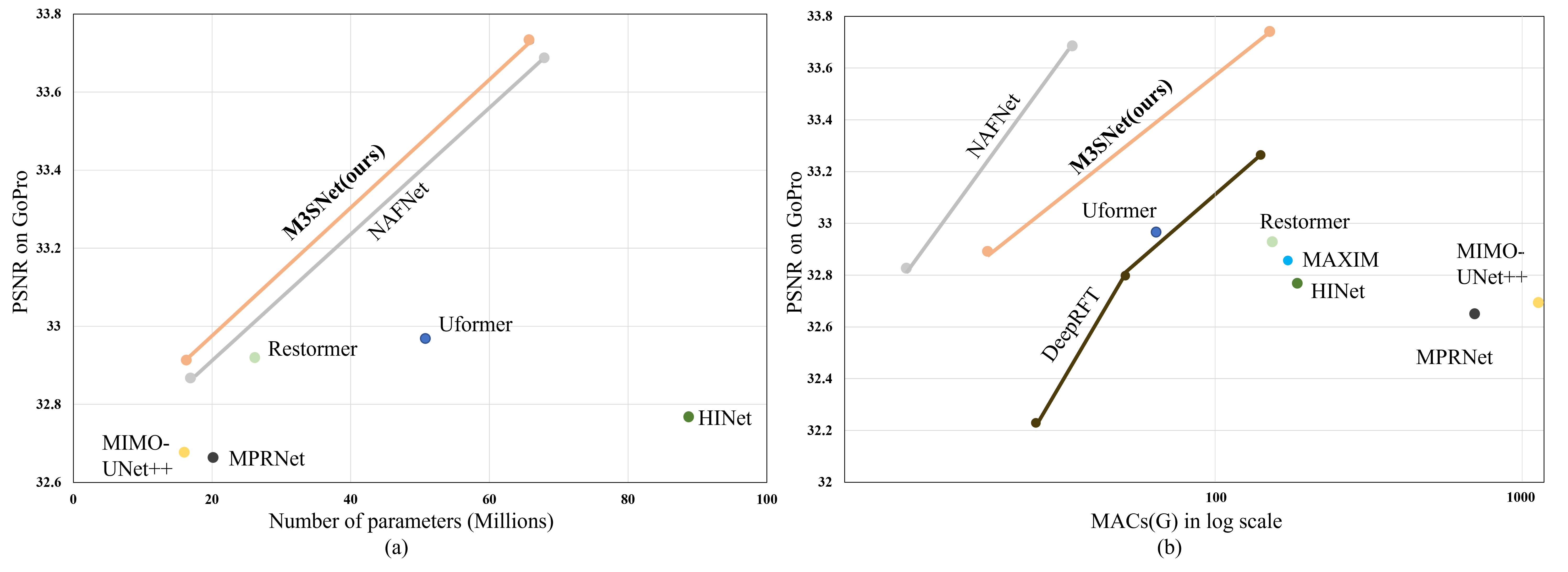}
	\caption{PSNR vs. computational cost on Image Deblurring. Under different parameter capacities, our model achieves state-of-the-art. In addition, our model involves only a relatively small number of multiply-accumulate operations (MACs).}
	\label{fig:01}
\end{figure}

The main contributions of this work are:
\begin{enumerate}
\item A novel single-stage approach capable of generating outputs that are contextually enriched and spatially accurate similar to a multi-stage architecture. Our architecture reduces system complexity due to its single-stage design eliminates the need for information to be passed between stages.
\item A feature fusion middleware mechanism that enables the exchange of information across multiple scales while preserving the fine details from the input image to the output image.
\item A multi-head attention middle block captures more global information. 
\item We demonstrate the versatility of M3SNet by setting new state-of-the-art on 6 synthetic and real-world datasets for various restoration tasks (image deraining and deblurring) while maintaining low complexity (see Fig.\ref{fig:01}). Further, we provide detailed analysis, qualitative results, and generalization tests.
\end {enumerate}

\section{Related Work}
Image degradation is a common occurrence caused by camera equipment and a variety of environmental factors. Depending on the specific degradation phenomenon, different image restoration tasks are proposed, e.g., deblurring and deraining. Early image restoration work was mainly based on manually crafting some prior knowledge, such as total variation and self-similarity~\cite{1992Nonlinear, 1997Prior, 2002Scale, 2005Fields, 2010Single, 2011Image, 2011Single}. With the rise of deep learning, data-driven methods like CNN~\cite{r7anwar2019drln,r105zhang2018rcan,r107zhang2020rdnir,r18dudhane2021burst,Zamir2021MPRNet,Zamir2022MIRNetv2} and Transformer~\cite{conde2022swin2sr,liang2021swinir,Zamir2021Restormer,Tsai2022Stripformer,Wang_2022_CVPR} have become the dominant approach for image restoration due to their impressive performance. These methods can be categorized as either single-stage or multi-stage based on their architectural design.

\subsection{Single-Stage Architecture.}
In recent years, the majority of image restoration research has focused on single-stage architecture. Among these architectures, the encoder-decoder based U-Net~\cite{chen2022simple, chu2022nafssr, Wang_2022_CVPR,2021Rethinking,deganv2,Zamir2021MPRNet,r90ECCV2020_984,r99Zhang2020PlugandPlayIR}  and dual network structure~\cite{2018DehazeGAN,2019Dense,2019Dual,tian2021designing,2020Refining} are mainly included.

\textbf{Encoder-Decoder Approaches.} In recent years, encoder-decoder have gained great attention from researchers in the field of image restoration thanks to its ability to capture multi-scale information. To construct an effective and efficient Transformer-based architecture for image restoration,~\cite{Wang_2022_CVPR} introduce a novel locally-enhanced window and multi-scale restoration modulator to create a hierarchical encoder-decoder network.~\cite{Zamir2022MIRNetv2} utilize selective kernel feature fusion to realize the information exchange of different scales and information aggregation based on attention.~\cite{9156921} develops a simple yet effective boosted decoder to progressively restore the haze-free image by incorporating the strengthen-operate-subtract boosting strategy in the decoder.  By eliminating or substituting the nonlinear activation function,~\cite{chen2022simple} establishes a simple baseline that yields measurable outcomes while requiring fewer computing resources. 

\textbf{Dual Network Approaches.} The Dual Networks architecture is designed with two parallel branches that separately estimate the structure and detail components of the target signals from the input. These components are then combined to reconstruct the final results according to the specific task formulation module. This architecture was first proposed by~\cite{2018LearningD} and has since inspired a lot of subsequent work, including in the areas of image dehazing~\cite{2018DehazeGAN,2019Dense,2019Dual}, image deraining~\cite{2018Fast}, image denoising~\cite{tian2021designing}, and image super-resolution/deblurring~\cite{2020Refining}. The Dual Networks approach has proven to be effective in addressing various low-level vision problems, by enabling a better separation of the structure and detail information, leading to improved performance in terms of both accuracy and computational efficiency. Furthermore, the flexibility of this architecture makes it adaptable to different types of data, making it a popular choice in the field of image restoration.

Despite the significant achievements made by these networks, it remains a challenge to effectively balance these competing goals of preserving spatial details and contextualized information while recovering images.
\subsection{Multi-Stage Architecture.}
The multi-stage networks are shown to be more effective than their single-stage counterparts in high level vision problems  ~\cite{Li2019RethinkingOM,15Cheng2019SPGNetSP,26Ghosh2018StackedSG,45li2020ms}. In recent years, there have been some attempts~\cite{2018Scale,2018Lightweight,RESCAN,PREnet,Zamir2021MPRNet,zhang2022event,Zhang_2019_CVPR} to apply multi-stage networks to image restoration. They aims to break down the image restoration process into several manageable stages, enabling the use of lightweight subnetworks to progressively restore clear images.  This approach facilitates the capture of both spatial details and contextualized information by individual subnetworks at each stage. To prevent the production of suboptimal results that may arise from using the same subnetwork at each stage, a supervisory attention mechanism was proposed along with the adoption of distinct subnetwork structures~\cite{Zamir2021MPRNet}. Additionally,~\cite{zhang2022event} present a novel self-supervised event-guided deep hierarchical Multi-patch Network to handle blurry images and videos through fine-to-coarse hierarchical localized representations.  Nevertheless, this approach elevates the complexity of the system as refining the previous stage's results is required in subsequent stages.

\section{Method}
\begin{figure}[tb] 
	\centering
	\includegraphics[width=1\textwidth]{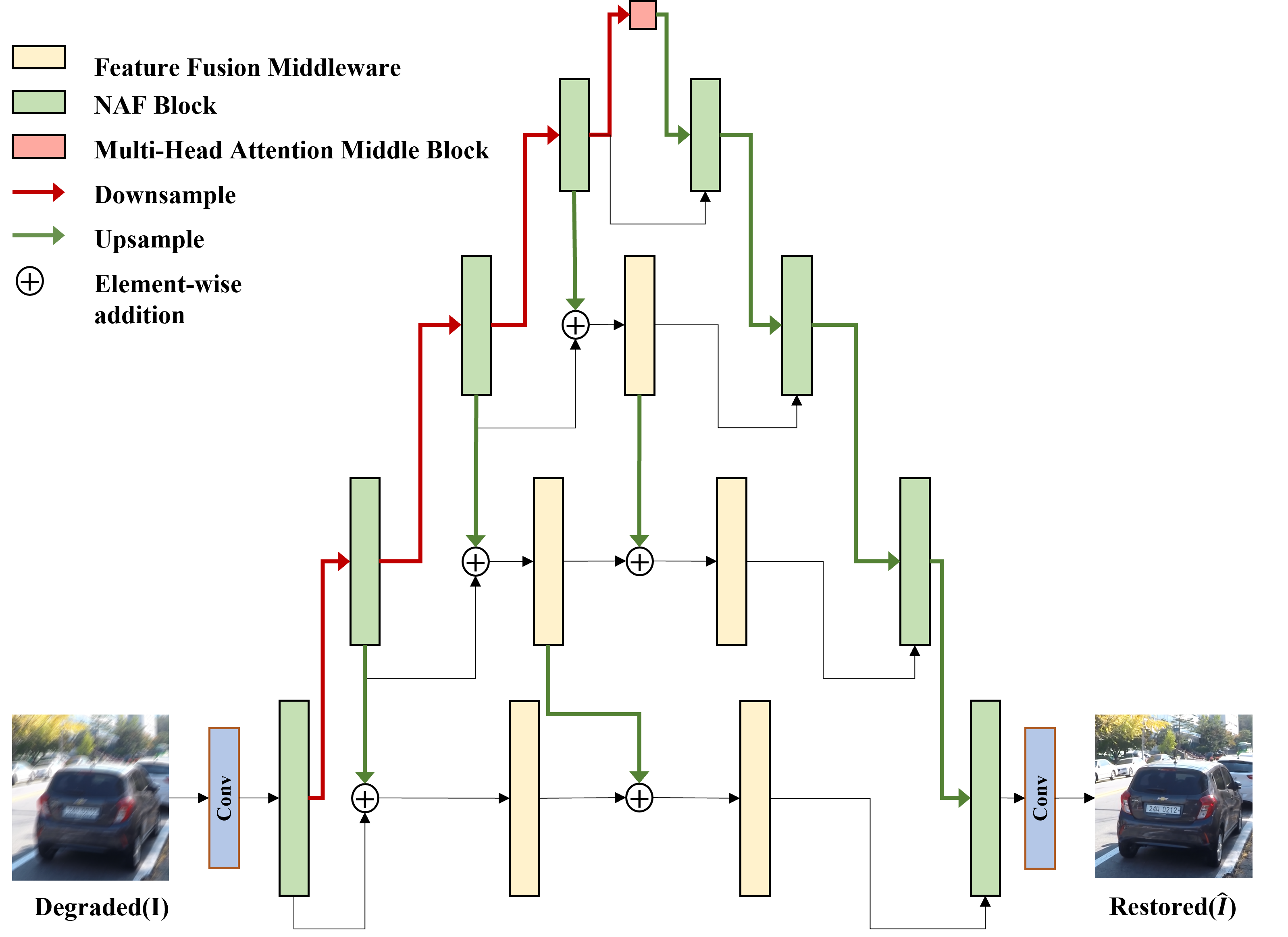}
	\caption{Architecture of M3SNet for image restoration.}
	\label{fig:02}
\end{figure}

Our primary goal is to create a single-stage network architecture that can efficiently handle the challenging task of image restoration by balancing the need for spatial details and context information, all while using fewer computational resources. The M3SNet is built upon a U-Net architecture, as shown in Figure~\ref{fig:02}. As is apparent from the figure, in contrast to the traditional U-Net network, we have inverted the architecture and introduced two key components: \textbf{(a)} the feature fusion middleware (FFM) and \textbf{(b)} the multi-head attention middle block (MHAMB). The model's architecture takes on a mountain-like shape, and we liken the image restoration process to climbing a mountain. 

\noindent\textbf{Overall Pipeline.} Given a degraded image $\mathbf{I} \in \mathbb R^{H \times W \times 3}$, M3SNet first applies a $3 \times 3$ convolutional layer to extract shallow feature maps $\mathbf{F_{0}} \in \mathbb R^{H \times W \times C}$ ($H, W, C$ are the feature map height, width, and channel number, respectively). Next these shallow features $\mathbf{F_{0}}$ pass through $4-level$ encoder-decoder and one multi-head attention middle block, yielding deep features $\mathbf{F_{DF}} \in \mathbb R^{H \times W \times C}$. Each layer contains multiple feature fusion middleware between the encoder and decoder to capture multi-scale information and retain spatial details. 
Finally, we apply convolution to deep  features $\mathbf{F_{DF}}$ and generate a residual image $\mathbf{R}\in  R^{H \times W \times 3}$ to which degraded image is added to obtain the restored image:$\mathbf{\hat{I}} = \mathbf{R} +\mathbf{I}$. We optimize the proposed network using PSNR loss : 
\begin{equation}
	\label{equ:01}
	PSNR = 10 \cdot log_{10} \cdot \frac{(2^n-1)^2}{||\mathbf{\hat{I}}-\mathbf{\dot I}||^2}
\end{equation}


\noindent where $\mathbf{\dot I}$ denotes the ground-truth image.
\begin{figure}[tb] 
	\centering
	\includegraphics[width=1\textwidth]{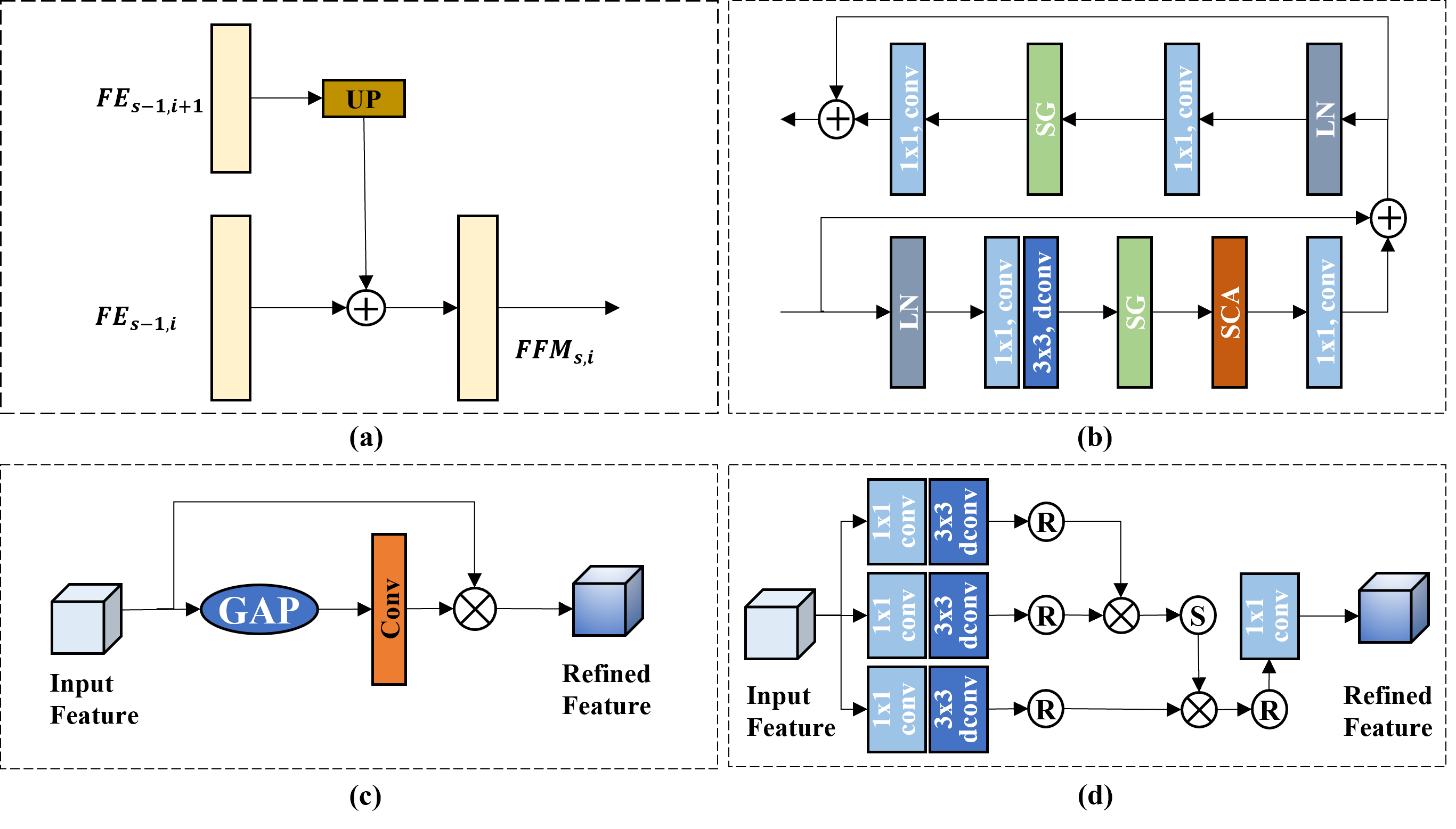}
	\caption{(a) Feature fusion middleware (FFM) that enables the exchange of information across multiple scales while preserving the fine details. (b) The architecture of nonlinear activation free block (NAFBlock). (c) Simplified Channel Attention (SCA). (d) Multi-head attention middle block (MHAMB) that  captures more global information.}
	\label{fig:03}
\end{figure}
\subsection{Feature Fusion Middleware (FFM)}
By incorporating an encoder and decoder network in the initial stage, followed by a network operating at the original image input resolution in the final stage, the multi-stage network can produce high-quality images with accurate spatial details and reliable contextual information. However, the latter stage of this process requires revising the results of the previous stage, which adds a slight level of complexity to the system. While a single-stage network has relatively less complexity, it may struggle to balance spatial details and context information effectively. Therefore, we are exploring a middleware mechanism for feature fusion that enables a simple single-stage U-Net architecture to achieve the same functionality as a multi-stage architecture, without sacrificing smoothness of expression or compromising on the original meaning.

As shown in Fig.~\ref{fig:03}(a), the feature fusion middleware(FFM) is a  nonlinear activation-free block (NAFBlock) with upsample and feature fusion. We have introduced the FFM between the encoder-decoder architectural levels to integrate upper-layer information into adjacent lower layers. This integration takes place sequentially, from the highest layer down to the lowest, until all information is fused into the original image resolution manipulation level. This approach enhances the network's capacity to capture and fuse multi-scale features, ranging from simple patterns at low levels, such as corner or edge/color connections, to more complex higher-level features, such as significant variations and specific objects. As a result, this structure preserves spatial details while integrating contextual information, ultimately ensuring high-quality image restoration. Formally, let  $\mathbf{FE_i} \in \mathbb R^{\frac{H}{i^2} \times \frac{W}{i^2} \times {i^2}C}$ be the output in the $i$-th level encoder $(i=1,2,3,4)$. At each level, the feature fusion information $FFM_{s,i}$ is given as:
\begin{equation}
	\label{equ:02}
	FFM_{1,i} =  H_{Naf_{1,i}}( FE_{i} \oplus UP(FE_{i+1}))
\end{equation}
\begin{equation}
	\label{equ:03}
	FFM_{s,i} =   H_{Naf_{s,i}}(UP(FFM_{s-1,i+1}) \oplus FFM_{s-1,i})
\end{equation}
where $\oplus$ denote the element-wise addition, $UP(\cdot)$ represents the up-sampling operation and $H_{Naf_{s,i}}(\cdot)$ is the $s$-th FFM in the $i$-th level. 

This design offers two benefits. Firstly, the feature fusion middleware integrates multi-scale information, allowing the network model to capture abundant context information. Secondly, the feature fusion middleware in the $1_{th}$ layer operates on the original image resolution, without employing any subsampling operation, thereby enabling the network model to acquire detailed spatial information of high-resolution features.

\noindent\textbf{NAFBlock.} NAFBlock is a variant of the U-Net network that simplifies the system by replacing or removing the nonlinear activation function. Fig.~\ref{fig:03}(b) illustrates the process of obtaining an output$Y$ from an input $X$ using Layer Normalization, Convolution, Simple Gate, and Simplified Channel Attention. Express as follows:
\begin{equation}
	\label{equ:04}
	X_1 = X + C_1(SCA(SG(C_3(C_1(LN(X))))))
\end{equation}
\begin{equation}
	\label{equ:05}
    Y = X_1 + C_1(SG(C_1(LN(X_1))))
\end{equation}
\begin{equation}
	\label{equ:06}
 SG = X_{f1} \times X_{f2}
\end{equation}
where $C_1$ is the $1 \times 1$ convolution, $C_3$ is the $3 \times 3$ depth-wise convolution, GAP is the adaptive average pool, $X_{f1}, X_{f2} \in \  R^{H \times W \times \frac{C}{2}}$ are obtained by dividing $X_{f3}$ into channel dimensions, and $SCA(\cdot)$ is shown in Fig.~\ref{fig:03}(c).

Finally, the depth features $\mathbf{F_{DF}}$ are obtained through this single-stage architecture, as demonstrated below:
\begin{equation}
	\label{equ:08}
    FD_{i} =  H_{Naf_{\textrm{-1},i}}(FD_{i+1} + FFM_{\textrm{-1},i}) 
\end{equation}
\begin{equation}
	\label{equ:09}
    F_{DF} = FD_{1}  
\end{equation}
where $FD_i$ is the output in the $i$-th level decoder, and -$1$ indicates that this is the last feature fusion middleware at this level.

\subsection{Multi-Head Attention Middle Block (MHAMB)}
The transformer model~\cite{Wang_2022_CVPR,Zamir2021Restormer,Tsai2022Stripformer} has gained popularity in image restoration tasks due to its capability to capture global information, as evidenced by its increasing usage in recent years. The computation on a global scale results in a quadratic complexity in relation to the number of tokens as shown in Eq.~\ref{equ:10}, rendering it inadequate for the representation of high-resolution images.
\begin{equation}
	\label{equ:10}
    \mathcal{O}_{MSA} = 4hwC^2 + 2(hw)^2C  
\end{equation}
To alleviate this issue,~\cite{Tsai2022Stripformer,Zamir2021Restormer,liang2021swinir} etc., proposed various methods to reduce complexity. In this paper, we propose a  multi-head attention middle block (MHAMB) as the bridge of the encoder-decoder, shown in Fig.~\ref{fig:03}(d).MHAMB utilizes global self-attention to process and integrate the feature map information that is generated by the last layer of the encoder. This approach is particularly efficient in handling large images because convolution downsamples space, while attention can effectively process smaller resolutions for better performance.
From the last layer of the encoder output $FE_4$, our MHAMB first apply a $1 \times 1$ convolution and a $3 \times 3$ depth-wise convolution to generate the $query, key$ and $value$ matrices $\mathbf{Q} \in \mathbb R^{H \times W \times C}, \mathbf{K} \in \mathbb R^{H \times W \times C}$ and $\mathbf{V} \in \mathbb R^{H \times W \times C}$ as follows:
\begin{equation}
\begin{aligned}
	\label{equ:11}
    &Q = C_3(C_1(FE_4))  \\
    &K = C_3(C_1(FE_4))  \\
    &V = C_3(C_1(FE_4))
\end{aligned}
\end{equation}
Then, we reshape $\mathbf{Q}$, $\mathbf{K}$, $\mathbf{V}$ to $\mathbf{\hat{Q}} \in \mathbb R^{(H W) \times \frac{C}{h} \times h}$, $\mathbf{\hat{K}} \in \mathbb R^{(H W) \times \frac{C}{h} \times h}$, $\mathbf{\hat{V}} \in \mathbb R^{(H W) \times \frac{C}{h} \times h}$, where $h$ is the number of head.
Next, the attention matrix is thus computed by the self-attention mechanism  as :
\begin{equation}
	\label{equ:10}
    Attention(\hat{Q}, \hat{K}, \hat{V}) =  SoftMax(\frac{\hat{Q}\hat{K}}{\beta})\hat{V} 
\end{equation}
where $\beta$ is a learning scaling parameter used to adjust the magnitude of the dot product of $\hat{Q}$ and $\hat{K}$ prior to the application of the softmax function. Finally, we reshape the attention matrix back to its original dimensions of $\mathbb R^{H \times W \times C}$ and apply a $1 \times 1$ convolution. The resulting output is then added to $FE_4$ and passed to the decoder. This allows us to capture more information and enhance the overall performance of the model as shown in the following experiment.

\section{Experiments}
\begin{figure} 
    \centerline{\includegraphics[width=1\textwidth]{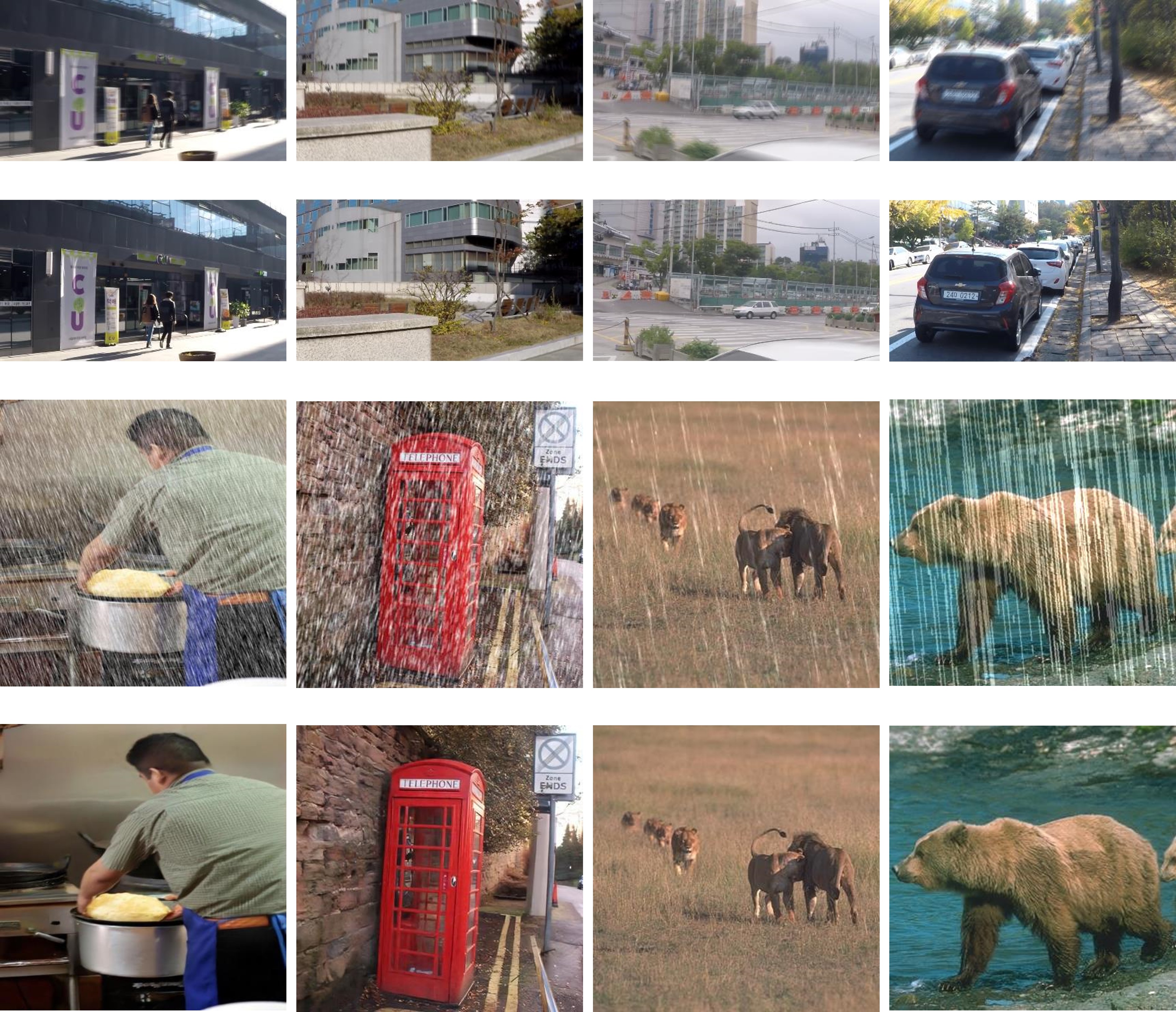}}
	\caption{Visualized results of M3SNet on various image restoration tasks. For each image pair, the upper one is degraded and the down one is predicted by M3SNet.
}
	\label{fig:05}
\end{figure}

We evaluate the proposed M3SNet on benchmark datasets  for two image restoration tasks, including \textbf{(a)} image deraining, and \textbf{(b)} image deblurring. Fig.~\ref{fig:05} displays some of the images predicted by our method. Our model recovered clearer images that were close to the ground truth.

\subsection{Datasets and  Evaluation Protocol}
We use PSNR and SSIM as quality assessment metrics. To report the reduction in error for each method relative to the best-performing method, we convert PSNR to RMSE ($RMSE \propto \sqrt{10^{-PSNR/10}}$) and SSIM to DSSIM (DSSIM = (1 - SSIM)/2).

\noindent\textbf{Image Deraining.} As shown in Table.~\ref{tb:000}. Our derain model is trained on a collection of 13,712 clean-rain image pairs obtained from multiple datasets~\cite{238099669,81Yang2016DeepJR,90Zhang2017ImageDU,DIDMDN}. We assess the model's performance on various test sets, including Test100~\cite{90Zhang2017ImageDU}, Test1200~\cite{DIDMDN}, Rain100H~\cite{81Yang2016DeepJR}, and Rain100L~\cite{81Yang2016DeepJR}.

\noindent\textbf{Image Deblurring.}As shown in Table.~\ref{tb:001}. To perform image deblurring, we utilize the GoPro~\cite{Gopro} dataset, which consists of 2,103 image pairs for training and 1,111 pairs for evaluation. Additionally, we assess the generalizability of our approach by applying the GoPro-trained model directly to the test images of the HIDE dataset. The HIDE dataset is designed specifically for human-aware motion deblurring, and its test set comprises 2,025 images.

\begin{table}
\caption{ Dataset description for image deraining.
}
\label{tb:000}
\begin{tabular}{cccc}
    \hline
     Datasets &Train Samples &Test Samples &Testset Rename
    \\
    \hline\hline
     Rain14000~\cite{238099669} &11200 &0 & -
     \\
    Rain1800~\cite{81Yang2016DeepJR} &1800 &0 & -
        \\
    Rain12~\cite{487780668} &12 &0 &- 
    \\
    Rain800~\cite{90Zhang2017ImageDU} &700 & 98 &Test100
    \\
    Rain1200~\cite{DIDMDN} &0 &1200 &Test1200
    \\
    Rain100H~\cite{81Yang2016DeepJR} &0 &100 & Rain100H
    \\
    Rain100L~\cite{81Yang2016DeepJR} &0 &100 &Rain100L
    \\
    \hline
\end{tabular}
\end{table}

\begin{table}
\caption{ Dataset description for image deblurring.
}
\label{tb:001}
\begin{tabular}{cccc}
    \hline
     Datasets &Train Samples &Test Samples &Testset Rename
    \\
    \hline\hline
    GoPro~\cite{Gopro} &2130 &1111 & GoPro
    \\
    HIDE~\cite{HIDE} &0 &2025 &HIDE
    \\
    \hline
\end{tabular}
\end{table}
\subsection{Implementation Details}
We train the proposed M3SNet without any pre-training and separate models for different image restoration tasks. We utilize the following block configurations in our network for each level: $[1,1,1,28]$ blocks for the encoder, $[1,1,1,1]$ blocks for the decoder, $[2,2,1,0]$ blocks for the FFM. And one MHAMB for the bridge of the encoder and decoder. 
We train models with Adam~\cite{2014Adam} optimizer($\beta_1=0.9, \beta_2=0.999$) and PSNR loss for $5 \times 10^5$ iterations with the initial learning rate $1 \times 10^{-3}$ gradually reduced to $1 \times 10^{-7}$  with the cosine annealing~\cite{2016SGDR}. We extract patches of size $256 \times 256$ from training images, and the batch size is set to $32$. We adopt TLC~\cite{Chu2021ImprovingIR} to solve the issue of performance degradation caused by training on patched images and testing on the full image. For data augmentation, we perform horizontal and vertical flips. 

\subsection{Image Deraining Results}
\begin{figure} 
    \centerline{\includegraphics[width=1\textwidth]{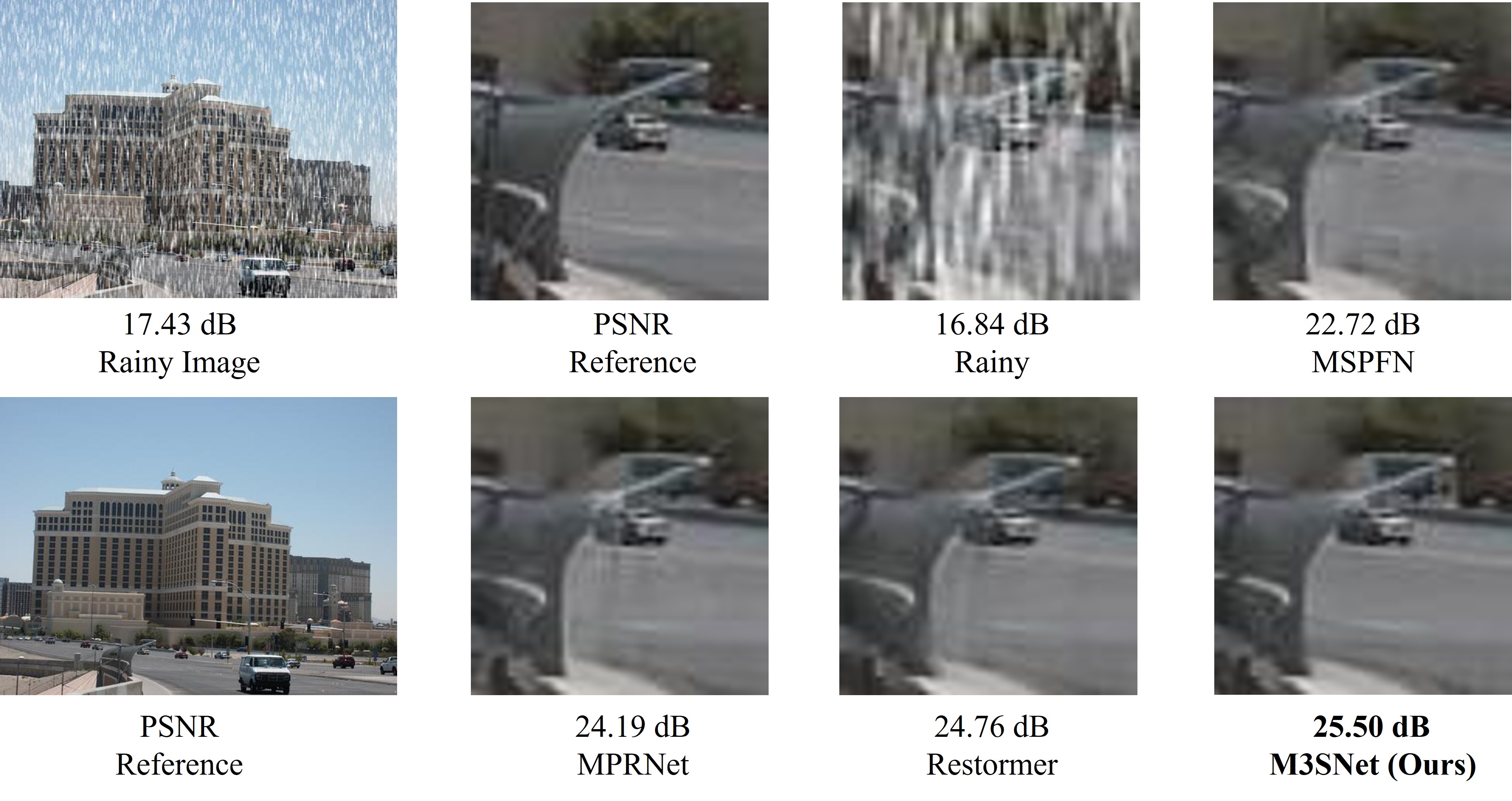}}
	\caption{\textbf{Image deraining} example. The outputs of M3SNet exhibit
no traces of rain streaks on both image samples. M3SNet also recovers the most detailed images.}
	\label{fig:06}
\end{figure}
In our image deraining task, we compute the PSNR/SSIM scores using the Y channel in the YCbCr color space, which is consistent with previous works such as\cite{Zamir2021MPRNet,MSPFN,SPAIR}. Our method has been demonstrated to outperform existing approaches significantly and consistently, as presented in Table~\ref{tb:01}. Specifically, our method achieves a remarkable improvement of 0.93 dB and a $10.2\%$ error reduction on average across all datasets when compared to the best CNN-based method, SPAIR~\cite{SPAIR}. Moreover, we can achieve up to 2.76 dB improvement over HINet~\cite{Chen_2021_CVPR} on individual datasets, such as Rain100L.

In addition to quantitative evaluations, Fig.~\ref{fig:06} presents qualitative results that demonstrate the effectiveness of our M3SNet in removing rain streaks of various orientations and magnitudes while preserving the structural content of the images.

\begin{sidewaystable}
\caption{Image deraining results. The best and second best scores are \textbf{highlighted} and \underline{underlined}. Our M3SNet is better than the state-of-the-art by 0.93 dB. }\label{tb:01}
\begin{tabular}{ccccccccc||cccc}
    \hline
    \multicolumn{1}{c}{} & \multicolumn{2}{c}{Test100~\cite{Test100}}  & \multicolumn{2}{c}{Test1200~\cite{MSPFN}} & \multicolumn{2}{c}{Rain100H~\cite{Rain100}} & \multicolumn{2}{c||}{Rain100L~\cite{Rain100}} & \multicolumn{4}{c}{Average} 
    \\
   Methods &PSNR $\uparrow$ &  SSIM $\uparrow$  & PSNR $\uparrow$ & SSIM $\uparrow$ &PSNR $\uparrow$ &SSIM $\uparrow$ & PSNR $\uparrow$&SSIM $\uparrow$ & \multicolumn{2}{c}{PSNR $\uparrow$} & \multicolumn{2}{c}{SSIM $\uparrow$}
    \\
    \hline\hline
    DerainNet~\cite{DerainNet}  & 22.77 & 0.810  & 23.38  & 0.835  & 14.92 &  0.592  & 27.03 & 0.884 & 22.48 & \textcolor{red}{(73.0\%)} & 0.796 & \textcolor{red}{(63.7\%)}
    \\
     SEMI~\cite{semi}  & 22.35  & 0.788  & 26.05 & 0.822  & 16.56 &  0.486 & 25.03 & 0.842 & 22.88 & \textcolor{red}{(71.7\%)} & 0.744 &\textcolor{red}{(71.1\%)}
     \\
    DIDMDN~\cite{DIDMDN} & 22.56 & 0.818  & 29.65 & 0.901  & 17.35 &  0.524 &25.23 & 0.741 & 24.58 & \textcolor{red}{(65.6\%)} & 0.770&\textcolor{red}{(67.8\%)}
     \\
    UMRL~\cite{UMRL}  & 24.41 &0.829 & 30.55 &  0.910   & 26.01 & 0.832 & 29.18 & 0.923& 28.02 & \textcolor{red}{(48.8\%)} & 0.880 & \textcolor{red}{(38.3\%)}
       \\
    RESCAN~\cite{RESCAN}  & 25.00 & 0.835 & 30.51 & 0.882  &26.36 & 0.786 & 29.80 &0.881 & 28.59 &\textcolor{red}{(45.4\%)} & 0.857&\textcolor{red}{(48.3\%)}
       \\
     PreNet~\cite{PREnet}  & 24.81 &0.851 & 31.36&  0.911   & 26.77 & 0.858  &32.44 & 0.950 &29.42 & \textcolor{red}{(39.9\%)} & 0.897 &\textcolor{red}{(28.2\%)}
    \\
   MSPFN~\cite{MSPFN}  & 27.50 & 0.876 & 32.39 &  0.916   & 28.66 & 0.860  & 32.40 & 0.933 & 30.75 &\textcolor{red}{(29.9\%)} & 0.903&\textcolor{red}{(23.7\%)}
       \\
     MPRNet~\cite{Zamir2021MPRNet}  & 30.27 & 0.897 & 32.91 &  0.916   & 30.41 & 0.890  & 36.40 & 0.965 & 32.73 &\textcolor{red}{(12.0\%)} & 0.921&\textcolor{red}{(6.3\%)}
       \\
     SPAIR~\cite{SPAIR}  & 30.35 & \underline{0.909} & 33.04 &  0.922   & \underline{30.95} & 0.892  & 36.93 & 0.969& 32.91 &\textcolor{red}{(10.2\%)}& \textbf{0.926}&\textcolor{red}{(0.0\%)}
     \\
      HINet~\cite{Chen_2021_CVPR}  & 30.29 & 0.906 & 33.05&  0.919   & 30.65 & \underline{0.894}  & 37.28 & 0.970 & 32.81 &\textcolor{red}{(11.2\%)} & 0.922&\textcolor{red}{(5.1\%)}
     \\
     \hline
      \textbf{M3SNet-32}  & \underline{31.29} & 0.903 & \underline{33.46}& 0.924   & 30.64 & 0.892  & \underline{39.62} & \underline{0.984} & \underline{33.75}   &\textcolor{red}{(1.0\%)} & \textbf{0.926}&\textcolor{red}{(0.0\%)}
      \\
      \textbf{M3SNet-64}  & 31.25 & 0.901 & \textbf{33.52} &  \underline{0.925}  & 30.54 & 0.889  & \textbf{40.04} & \textbf{0.985} & \textbf{33.84} &\textcolor{red}{(0.0\%)} & \underline{0.925}&\textcolor{red}{(1.3\%)}
    \\
    \hline
\end{tabular}
\end{sidewaystable}

\subsection{Image Deblurring Result}

\begin{figure} 
    \centerline{\includegraphics[width=1\textwidth]{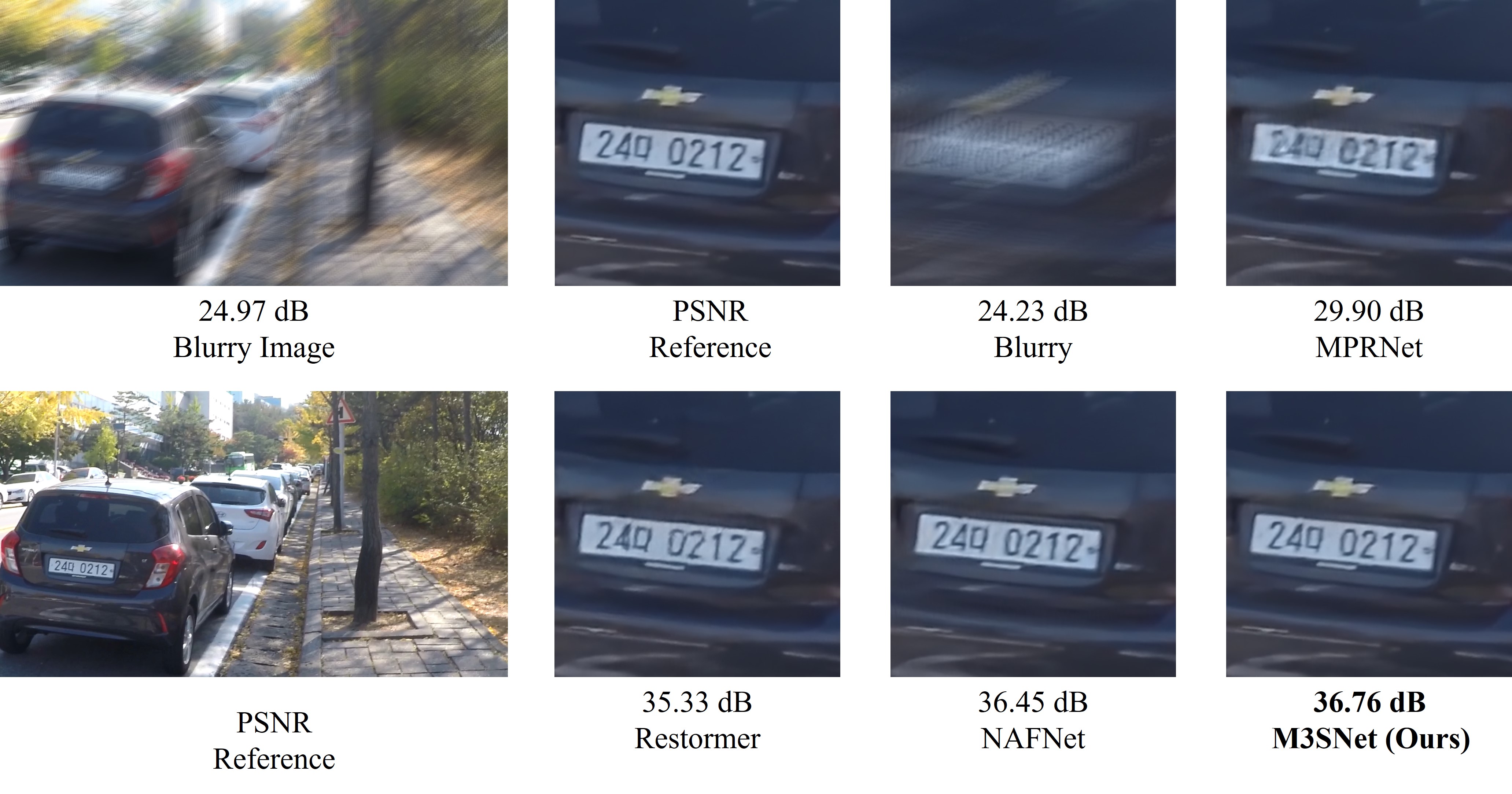}}
	\caption{\textbf{Image deblurring} example on the GoPro dataset~\cite{Gopro}. Compared to the state-of-the-art methods, our M3SNet
restores sharper and perceptually-faithful images.}
	\label{fig:07}
\end{figure}

The performance evaluation of image deblurring approaches on the GoPro~\cite{Gopro} and HIDE~\cite{HIDE} datasets is presented in Table~\ref{tb:02}. Our M3SNet outperformed other methods, with a performance gain of 0.09dB when averaging across all datasets~\cite{Gopro,HIDE}. Compared to our baseline network NAFNet~\cite{chen2022simple}, we improve 0.04 dB and 0.05 dB at 32 widths and 64 widths, respectively. It is worth noting that even though our network is trained solely on the GoPro Dataset, it still achieves state-of-the-art results (31.49 dB in PSNR) on the HIDE dataset. This demonstrates its impressive generalization capability.

Figure~\ref{fig:07} displays some of the deblurred images produced by our  method. Our model recovered clearer images that were closer to the ground truth than those by others.

\begin{table}[ht]
\caption{Image deblurring results. The proposed M3SNet is trained only on the GoPro dataset but achieves a 0.09 dB improvement over the state of the art on the average of the effects on both datasets. \label{tb:02}}

\begin{tabular}{ccccc||cccc}
    \hline
    \multicolumn{1}{c}{} & \multicolumn{2}{c}{GoPro~\cite{Gopro}}  & \multicolumn{2}{c||}{HIDE~\cite{HIDE}} & \multicolumn{4}{c}{Average}
    \\
   Methods & PSNR $\uparrow$ & SSIM $\uparrow$ & PSNR $\uparrow$ & SSIM $\uparrow$   & \multicolumn{2}{c}{ PSNR $\uparrow$} & \multicolumn{2}{c}{ SSIM $\uparrow$} 
    \\
    \hline\hline
    DeblurGAN~\cite{Degan}  & $28.70 $ & $0.858$ & $24.51$ & $0.871$& 26.61 &\textcolor{red}{(49.9\%)} &0.865 &\textcolor{red}{(70.4\%)}
    \\
    Nah $et al.$~\cite{Gopro} & $29.08$ & $0.914$ & $25.73$ & $0.874$& 27.41  &\textcolor{red}{(45.1\%)}  &0.894&\textcolor{red}{(62.3\%)}
    \\
    DeblurGAN-v2~\cite{deganv2} & 29.55 & 0.934 & 26.61 & 0.875& 28.08  &\textcolor{red}{(40.7\%)} &0.905&\textcolor{red}{(57.9\%)}
    \\
    SRN~\cite{2018Scale} & 30.26 & 0.934 & 28.36 & 0.915 &29.31  &\textcolor{red}{(31.7\%)} &0.925&\textcolor{red}{(46.7\%)}
    \\
    Gao $et al.$~\cite{Gao2019DynamicSD} & 30.90 & 0.935 & 29.11 & 0.913 &30.01  &\textcolor{red}{(26.0\%)} &0.924&\textcolor{red}{(47.4\%)}
    \\
    DBGAN~\cite{DBGAN} & 31.10 & 0.942 & 28.94 & 0.915 &30.02 &\textcolor{red}{(25.9\%)} &0.929&\textcolor{red}{(43.7\%)}
    \\
    MT-RNN~\cite{MTRNN} & 31.15 & 0.945 & 29.15 & 0.918 &30.15  &\textcolor{red}{(24.8\%)}&0.932&\textcolor{red}{(41.2\%)}
    \\
    DMPHN~\cite{Zhang_2019_CVPR} & 31.20 & 0.940 & 29.09 & 0.924 & 30.15  &\textcolor{red}{(24.8\%)}&0.932&\textcolor{red}{(41.2\%)}
    \\
    Suin $et al.$~\cite{Suin2020SpatiallyAttentivePN} & 31.85 & 0.948 & 29.98 & 0.930&30.92 &\textcolor{red}{(17.8\%)}&0.939&\textcolor{red}{(34.4\%)}
    \\
    SPAIR~\cite{SPAIR} & 32.06 & 0.953 & 30.29 & 0.931 &31.18  &\textcolor{red}{(15.3\%)}&0.942&\textcolor{red}{(21.0\%)}
    \\
    MIMO-UNet++~\cite{2021Rethinking} & 32.45 & 0.957 & 29.99 & 0.930 &31.22  &\textcolor{red}{(14.9\%)}&0.944&\textcolor{red}{(28.6\%)}
    \\
    MPRNet~\cite{Zamir2021MPRNet} & 32.66 & 0.959 & 30.96 & 0.939 &31.81 &\textcolor{red}{(8.9\%)} &0.949&\textcolor{red}{(21.6\%)}
    \\
    MPRNet-local~\cite{Zamir2021MPRNet} & 33.31 & 0.964 &31.19 &0.945 &32.25 &\textcolor{red}{(4.2\%)}&0.955&\textcolor{red}{(11.1\%)}
    \\
    Restormer~\cite{Zamir2021Restormer} & 32.92 & 0.961 & \underline{31.22} & 0.942 &32.07 &\textcolor{red}{(6.1\%)}&0.952&\textcolor{red}{(16.7\%)}
    \\
    Restormer-local~\cite{Zamir2021Restormer} & 33.57 & \underline{0.966} & \textbf{31.49} & 0.945 &\underline{32.53}  &\textcolor{red}{(1.0\%)}&0.956&\textcolor{red}{(9.1\%)}
    \\
    Uformer~\cite{Wang_2022_CVPR} &32.97 & \textbf{0.967} &30.83 &\textbf{0.952} &31.90  &\textcolor{red}{(8.0\%)}&\textbf{0.960}&\textcolor{red}{(0.0\%)}
     \\
    HINet~\cite{Chen_2021_CVPR}&32.71&-&-&-&-&-
    \\
    HINet-local~\cite{Chen_2021_CVPR}&33.08&0.962&-&-&-&-
    \\
    \hline
    NAFNet-32~\cite{chen2022simple}&32.87&0.961&-&-&-&-
    \\
    NAFNet-64~\cite{chen2022simple}&\underline{33.69}&\textbf{0.967}&-&-&-&-
    \\
    \hline
    \textbf{M3SNet-32 (ours)}&32.91&0.965&30.92&0.948&31.92 &\textcolor{red}{(7.8\%)} &0.957&\textcolor{red}{(6.9\%)}
    \\
    \textbf{M3SNet-64 (ours)}&\textbf{33.74}&\textbf{0.967}&\textbf{31.49}&\underline{0.951}&\textbf{32.62} &\textcolor{red}{(0.0\%)} & \underline{0.959}&\textcolor{red}{(2.4\%)}
    \\
    \hline
\end{tabular}

\end{table}

\subsection{Ablation Studies}
The ablation studies are conducted on image deblurring (GoPro~\cite{Gopro}) to analyze the impact of each of our model components. It is worth noting that we use NAFNet-32~\cite{chen2022simple} as the baseline network to demonstrate the effectiveness by adding our proposed component. Table.~\ref{tb:03} shows the effectiveness of our M3SNet. Next, we describe the impact of each component.

\textbf{FFM}
We demonstrate the effectiveness of the proposed feature fusion middleware by adding it to NAFNet-32~\cite{chen2022simple}. As the Table.~\ref{tb:03} shows that the FFM added in each level of encoder-decoder leads to improved performance (32.90 dB) as compared to the baseline. This indicates that our proposed FFM can help the model capture more multi-scale information while also preserving spatial detail information.

\textbf{MHAMB} As the Table.~\ref{tb:03} shows a tiny improvement in PSNR from 30.87 dB to 30.88 dB after we add a multi-head attention middle block (MHAMB) to capture more global information in the topmost level. 

Adding both of these components improves the performance by a large margin from 32.87 dB to 32.91 dB in PSNR.

\begin{table}

\caption{Ablation study on individual components of the
proposed M3SNet. \label{tb:03}}
\begin{tabular}{ccccc}
\hline  
\multicolumn{1}{c}{ } & \multicolumn{1}{c}{NAFNet-32} & \multicolumn{1}{c}{+ FFM} & \multicolumn{1}{c}{+ MHAMB} & \multicolumn{1}{c}{+ FFM \& MHAMB}
  \\
\hline\hline
\multicolumn{1}{c}{PSNR} & \multicolumn{1}{c}{32.87} & \multicolumn{1}{c}{32.90} & \multicolumn{1}{c}{32.88} & \multicolumn{1}{c}{32.91}
\\
\multicolumn{1}{c}{SSIM} & \multicolumn{1}{c}{0.9606} & \multicolumn{1}{c}{0.9632} & \multicolumn{1}{c}{0.9606} & \multicolumn{1}{c}{0.9644}
\\
\hline
\end{tabular}

\end{table}

\subsection{Resource Efficient}

\begin{table}
\caption{The evaluation of model computational complexity. This is conducted with an input size of $256 \times 256$, on an NVIDIA 1060 GPU. \label{tb:04}}

\begin{tabular}{ccccc}
\hline
\multicolumn{2}{c}{Method} & \multicolumn{1}{c}{PSNR} & \multicolumn{1}{c}{Params(M)} & \multicolumn{1}{c}{MACs(G)} 
\\
\hline\hline
\multicolumn{2}{c}{MIMO-UNet++~\cite{2021Rethinking}}  & \multicolumn{1}{c}{$32.68$} & \multicolumn{1}{c}{$16.1$}  & \multicolumn{1}{c}{1235} 
\\
\multicolumn{2}{c}{MPRNet~\cite{Zamir2021MPRNet}}  & \multicolumn{1}{c}{$32.66$} & \multicolumn{1}{c}{20.1}  & \multicolumn{1}{c}{778.2}  
\\
\multicolumn{2}{c}{MPRNet-local~\cite{Zamir2021MPRNet}}  & \multicolumn{1}{c}{$33.31$} & \multicolumn{1}{c}{20.1}  & \multicolumn{1}{c}{778.2}  
\\
\multicolumn{2}{c}{HINet~\cite{Chen_2021_CVPR}}  & \multicolumn{1}{c}{$32.77$} & \multicolumn{1}{c}{88.7}  & \multicolumn{1}{c}{170.7} 
\\
\multicolumn{2}{c}{Restormer~\cite{Zamir2021Restormer}}  & \multicolumn{1}{c}{$32.92$} & \multicolumn{1}{c}{26.13}  & \multicolumn{1}{c}{140} 
\\
\multicolumn{2}{c}{Restormer-local~\cite{Zamir2021Restormer}}  & \multicolumn{1}{c}{$33.57$} & \multicolumn{1}{c}{26.13}  & \multicolumn{1}{c}{140} 
\\
\multicolumn{2}{c}{Uformer~\cite{Wang_2022_CVPR}}  & \multicolumn{1}{c}{$32.97$} & \multicolumn{1}{c}{50.88}  & \multicolumn{1}{c}{89.5} 
\\
\hline
\multicolumn{2}{c}{NAFNet-32~\cite{chen2022simple}}  & \multicolumn{1}{c}{$32.87$} & \multicolumn{1}{c}{17.1}  & \multicolumn{1}{c}{32}
\\
\multicolumn{2}{c}{NAFNet-64~\cite{chen2022simple}}  & \multicolumn{1}{c}{$33.69$} & \multicolumn{1}{c}{68}  & \multicolumn{1}{c}{65}
\\
\hline
\multicolumn{2}{c}{\textbf{M3SNet-32 (ours)}}  & \multicolumn{1}{c}{$32.91$} & \multicolumn{1}{c}{16.7}  & \multicolumn{1}{c}{37}
\\
\multicolumn{2}{c}{\textbf{M3SNet-64 (ours)}}  & \multicolumn{1}{c}{$33.74$} & \multicolumn{1}{c}{66.3}  & \multicolumn{1}{c}{146}
\\
\hline
\end{tabular}
\end{table}
Deep learning models have become increasingly complex in order to achieve higher accuracy. However, larger models require more resources and may not be practical in certain contexts. Therefore, there is a need to design lightweight image restoration models that can achieve high accuracy. In our work, we design a mountain-shaped single-stage network. This architecture optimizes the balance between spatial details and contextual information while minimizing the computational resources required to restore images.

Our M3SNet has been shown to outperform other models, as demonstrated in Table~\ref{tb:04}. Despite having 0.6M higher parameters than MIMO-unet++ \cite{2021Rethinking}, our proposed M3SNet-32 still achieves better performance, while using significantly fewer computational resources, with MACs  approximately 40 times smaller than that of MIMO-unet++. Although the MACs  of our model are higher than Uformer \cite{Wang_2022_CVPR} and NAFNet \cite{chen2022simple}, our model parameters are smaller, yet still perform better. Considering all factors, including model parameters, MACs, and performance, our model is the optimal choice.

\section{Conclusion}
In this paper, we present a single-stage network with a mountain-shaped structure that effectively captures multi-scale feature information and minimizes the computational resources required for image restoration. Our design is guided by the principle of balancing the competing goals of contextual information and spatial details while recovering images. To this end, we propose a feature fusion middleware mechanism that enables seamless information exchange between the encoder-decoder architecture's different levels. This approach smoothly combines upper-layer information with adjacent lower-layer information and eventually integrates all information to the original image resolution manipulation level. 
To overcome the limitations of CNNs' receptive fields and capture more global information, we utilize a multi-head attention middle block as the bridge of our encoder-decoder architecture.  Furthermore, to maintain computational efficiency and lightweight model size, we replace or remove nonlinear activation functions and instead use multiplication.  Our extensive experiments on multiple benchmark datasets demonstrate that our M3SNet model significantly outperforms existing methods while utilizing low computational resources.


\bibliography{sn-bibliography}


\begin{thebibliography}{70}
\ifx \bisbn   \undefined \def \bisbn  #1{ISBN #1}\fi
\ifx \binits  \undefined \def \binits#1{#1}\fi
\ifx \bauthor  \undefined \def \bauthor#1{#1}\fi
\ifx \batitle  \undefined \def \batitle#1{#1}\fi
\ifx \bjtitle  \undefined \def \bjtitle#1{#1}\fi
\ifx \bvolume  \undefined \def \bvolume#1{\textbf{#1}}\fi
\ifx \byear  \undefined \def \byear#1{#1}\fi
\ifx \bissue  \undefined \def \bissue#1{#1}\fi
\ifx \bfpage  \undefined \def \bfpage#1{#1}\fi
\ifx \blpage  \undefined \def \blpage #1{#1}\fi
\ifx \burl  \undefined \def \burl#1{\textsf{#1}}\fi
\ifx \doiurl  \undefined \def \doiurl#1{\url{https://doi.org/#1}}\fi
\ifx \betal  \undefined \def \betal{\textit{et al.}}\fi
\ifx \binstitute  \undefined \def \binstitute#1{#1}\fi
\ifx \binstitutionaled  \undefined \def \binstitutionaled#1{#1}\fi
\ifx \bctitle  \undefined \def \bctitle#1{#1}\fi
\ifx \beditor  \undefined \def \beditor#1{#1}\fi
\ifx \bpublisher  \undefined \def \bpublisher#1{#1}\fi
\ifx \bbtitle  \undefined \def \bbtitle#1{#1}\fi
\ifx \bedition  \undefined \def \bedition#1{#1}\fi
\ifx \bseriesno  \undefined \def \bseriesno#1{#1}\fi
\ifx \blocation  \undefined \def \blocation#1{#1}\fi
\ifx \bsertitle  \undefined \def \bsertitle#1{#1}\fi
\ifx \bsnm \undefined \def \bsnm#1{#1}\fi
\ifx \bsuffix \undefined \def \bsuffix#1{#1}\fi
\ifx \bparticle \undefined \def \bparticle#1{#1}\fi
\ifx \barticle \undefined \def \barticle#1{#1}\fi
\bibcommenthead
\ifx \bconfdate \undefined \def \bconfdate #1{#1}\fi
\ifx \botherref \undefined \def \botherref #1{#1}\fi
\ifx \url \undefined \def \url#1{\textsf{#1}}\fi
\ifx \bchapter \undefined \def \bchapter#1{#1}\fi
\ifx \bbook \undefined \def \bbook#1{#1}\fi
\ifx \bcomment \undefined \def \bcomment#1{#1}\fi
\ifx \oauthor \undefined \def \oauthor#1{#1}\fi
\ifx \citeauthoryear \undefined \def \citeauthoryear#1{#1}\fi
\ifx \endbibitem  \undefined \def \endbibitem {}\fi
\ifx \bconflocation  \undefined \def \bconflocation#1{#1}\fi
\ifx \arxivurl  \undefined \def \arxivurl#1{\textsf{#1}}\fi
\csname PreBibitemsHook\endcsname

\bibitem[\protect\citeauthoryear{Rudin et~al.}{1992}]{1992Nonlinear}
\begin{botherref}
\oauthor{\bsnm{Rudin}, \binits{L.I.}},
\oauthor{\bsnm{Osher}, \binits{S.}},
\oauthor{\bsnm{Fatemi}, \binits{E.}}:
Nonlinear total variation based noise removal algorithms.
Physica D Nonlinear Phenomena
(1992)
\end{botherref}
\endbibitem

\bibitem[\protect\citeauthoryear{Song and Mumford}{1997}]{1997Prior}
\begin{barticle}
\bauthor{\bsnm{Song}, \binits{C.Z.}},
\bauthor{\bsnm{Mumford}, \binits{D.}}:
\batitle{Prior learning and gibbs reaction-diffusion}.
\bjtitle{TPAMI}
\bvolume{19}(\bissue{11}),
\bfpage{1236}--\blpage{1250}
(\byear{1997})
\end{barticle}
\endbibitem

\bibitem[\protect\citeauthoryear{Perona and Malik}{2002}]{2002Scale}
\begin{barticle}
\bauthor{\bsnm{Perona}, \binits{P.}},
\bauthor{\bsnm{Malik}, \binits{J.}}:
\batitle{Scale-space and edge detection using anisotropic diffusion}.
\bjtitle{TPAMI}
\bvolume{12}(\bissue{7}),
\bfpage{629}--\blpage{639}
(\byear{2002})
\end{barticle}
\endbibitem

\bibitem[\protect\citeauthoryear{Roth and Black}{2005}]{2005Fields}
\begin{bchapter}
\bauthor{\bsnm{Roth}, \binits{S.}},
\bauthor{\bsnm{Black}, \binits{M.J.}}:
\bctitle{Fields of experts: A framework for learning image priors}.
In: \bbtitle{CVPR}
(\byear{2005})
\end{bchapter}
\endbibitem

\bibitem[\protect\citeauthoryear{Kim and Kwon}{2010}]{2010Single}
\begin{barticle}
\bauthor{\bsnm{Kim}, \binits{K.I.}},
\bauthor{\bsnm{Kwon}, \binits{Y.}}:
\batitle{Single-image super-resolution using sparse regression and natural
  image prior}.
\bjtitle{TPAMI}
\bvolume{32}(\bissue{6}),
\bfpage{1127}
(\byear{2010})
\end{barticle}
\endbibitem

\bibitem[\protect\citeauthoryear{Dong et~al.}{2011}]{2011Image}
\begin{barticle}
\bauthor{\bsnm{Dong}, \binits{W.}},
\bauthor{\bsnm{Zhang}, \binits{L.}},
\bauthor{\bsnm{Shi}, \binits{G.}},
\bauthor{\bsnm{Wu}, \binits{X.}}:
\batitle{Image deblurring and super-resolution by adaptive sparse domain
  selection and adaptive regularization}.
\bjtitle{TIP}
\bvolume{20}(\bissue{7}),
\bfpage{1838}--\blpage{1857}
(\byear{2011})
\end{barticle}
\endbibitem

\bibitem[\protect\citeauthoryear{He et~al.}{2011}]{2011Single}
\begin{botherref}
\oauthor{\bsnm{He}, \binits{K.}},
\oauthor{\bsnm{Sun}, \binits{J.}},
\oauthor{\bsnm{Tang}, \binits{X.}}:
Single image haze removal using dark channel prior.
TPAMI
(2011)
\end{botherref}
\endbibitem

\bibitem[\protect\citeauthoryear{Zamir et~al.}{2021}]{Zamir2021MPRNet}
\begin{bchapter}
\bauthor{\bsnm{Zamir}, \binits{S.W.}},
\bauthor{\bsnm{Arora}, \binits{A.}},
\bauthor{\bsnm{Khan}, \binits{S.}},
\bauthor{\bsnm{Hayat}, \binits{M.}},
\bauthor{\bsnm{Khan}, \binits{F.S.}},
\bauthor{\bsnm{Yang}, \binits{M.-H.}},
\bauthor{\bsnm{Shao}, \binits{L.}}:
\bctitle{Multi-stage progressive image restoration}.
In: \bbtitle{CVPR}
(\byear{2021})
\end{bchapter}
\endbibitem

\bibitem[\protect\citeauthoryear{Chen et~al.}{2021}]{Chen_2021_CVPR}
\begin{bchapter}
\bauthor{\bsnm{Chen}, \binits{L.}},
\bauthor{\bsnm{Lu}, \binits{X.}},
\bauthor{\bsnm{Zhang}, \binits{J.}},
\bauthor{\bsnm{Chu}, \binits{X.}},
\bauthor{\bsnm{Chen}, \binits{C.}}:
\bctitle{Hinet: Half instance normalization network for image restoration}.
In: \bbtitle{CVPR}
(\byear{2021})
\end{bchapter}
\endbibitem

\bibitem[\protect\citeauthoryear{Ren et~al.}{2019}]{PREnet}
\begin{botherref}
\oauthor{\bsnm{Ren}, \binits{D.}},
\oauthor{\bsnm{Zuo}, \binits{W.}},
\oauthor{\bsnm{Hu}, \binits{Q.}},
\oauthor{\bsnm{Zhu}, \binits{P.F.}},
\oauthor{\bsnm{Meng}, \binits{D.}}:
Progressive image deraining networks: A better and simpler baseline.
2019 IEEE/CVF Conference on Computer Vision and Pattern Recognition (CVPR),
3932--3941
(2019)
\end{botherref}
\endbibitem

\bibitem[\protect\citeauthoryear{Li et~al.}{2018}]{RESCAN}
\begin{bchapter}
\bauthor{\bsnm{Li}, \binits{X.}},
\bauthor{\bsnm{Wu}, \binits{J.}},
\bauthor{\bsnm{Lin}, \binits{Z.}},
\bauthor{\bsnm{Liu}, \binits{H.}},
\bauthor{\bsnm{Zha}, \binits{H.}}:
\bctitle{Recurrent squeeze-and-excitation context aggregation net for single
  image deraining}.
In: \bbtitle{European Conference on Computer Vision}
(\byear{2018})
\end{bchapter}
\endbibitem

\bibitem[\protect\citeauthoryear{Chen et~al.}{2022}]{chen2022simple}
\begin{botherref}
\oauthor{\bsnm{Chen}, \binits{L.}},
\oauthor{\bsnm{Chu}, \binits{X.}},
\oauthor{\bsnm{Zhang}, \binits{X.}},
\oauthor{\bsnm{Sun}, \binits{J.}}:
Simple baselines for image restoration.
arXiv preprint arXiv:2204.04676
(2022)
\end{botherref}
\endbibitem

\bibitem[\protect\citeauthoryear{Chu et~al.}{2022}]{chu2022nafssr}
\begin{bchapter}
\bauthor{\bsnm{Chu}, \binits{X.}},
\bauthor{\bsnm{Chen}, \binits{L.}},
\bauthor{\bsnm{Yu}, \binits{W.}}:
\bctitle{Nafssr: Stereo image super-resolution using nafnet}.
In: \bbtitle{CVPR}
(\byear{2022})
\end{bchapter}
\endbibitem

\bibitem[\protect\citeauthoryear{Pan et~al.}{2022}]{2022Learning}
\begin{botherref}
\oauthor{\bsnm{Pan}, \binits{J.}},
\oauthor{\bsnm{Sun}, \binits{D.}},
\oauthor{\bsnm{Zhang}, \binits{J.}},
\oauthor{\bsnm{Tang}, \binits{J.}},
\oauthor{\bsnm{Yang}, \binits{J.}},
\oauthor{\bsnm{Tai}, \binits{Y.W.}},
\oauthor{\bsnm{Yang}, \binits{M.H.}}:
Dual convolutional neural networks for low-level vision.
IJCV
(2022)
\end{botherref}
\endbibitem

\bibitem[\protect\citeauthoryear{Zhang et~al.}{2023}]{zhang2023accurate}
\begin{bchapter}
\bauthor{\bsnm{Zhang}, \binits{J.}},
\bauthor{\bsnm{Zhang}, \binits{Y.}},
\bauthor{\bsnm{Gu}, \binits{J.}},
\bauthor{\bsnm{Zhang}, \binits{Y.}},
\bauthor{\bsnm{Kong}, \binits{L.}},
\bauthor{\bsnm{Yuan}, \binits{X.}}:
\bctitle{Accurate image restoration with attention retractable transformer}.
In: \bbtitle{ICLR}
(\byear{2023})
\end{bchapter}
\endbibitem

\bibitem[\protect\citeauthoryear{Zamir et~al.}{2022}]{Zamir2021Restormer}
\begin{bchapter}
\bauthor{\bsnm{Zamir}, \binits{S.W.}},
\bauthor{\bsnm{Arora}, \binits{A.}},
\bauthor{\bsnm{Khan}, \binits{S.}},
\bauthor{\bsnm{Hayat}, \binits{M.}},
\bauthor{\bsnm{Khan}, \binits{F.S.}},
\bauthor{\bsnm{Yang}, \binits{M.-H.}}:
\bctitle{Restormer: Efficient transformer for high-resolution image
  restoration}.
In: \bbtitle{CVPR}
(\byear{2022})
\end{bchapter}
\endbibitem

\bibitem[\protect\citeauthoryear{Tsai et~al.}{2022}]{Tsai2022Stripformer}
\begin{bchapter}
\bauthor{\bsnm{Tsai}, \binits{F.-J.}},
\bauthor{\bsnm{Peng}, \binits{Y.-T.}},
\bauthor{\bsnm{Lin}, \binits{Y.-Y.}},
\bauthor{\bsnm{Tsai}, \binits{C.-C.}},
\bauthor{\bsnm{Lin}, \binits{C.-W.}}:
\bctitle{Stripformer: Strip transformer for fast image deblurring}.
In: \bbtitle{ECCV}
(\byear{2022})
\end{bchapter}
\endbibitem

\bibitem[\protect\citeauthoryear{Wang et~al.}{2022}]{Wang_2022_CVPR}
\begin{bchapter}
\bauthor{\bsnm{Wang}, \binits{Z.}},
\bauthor{\bsnm{Cun}, \binits{X.}},
\bauthor{\bsnm{Bao}, \binits{J.}},
\bauthor{\bsnm{Zhou}, \binits{W.}},
\bauthor{\bsnm{Liu}, \binits{J.}},
\bauthor{\bsnm{Li}, \binits{H.}}:
\bctitle{Uformer: A general u-shaped transformer for image restoration}.
In: \bbtitle{CVPR}
(\byear{2022})
\end{bchapter}
\endbibitem

\bibitem[\protect\citeauthoryear{Wang et~al.}{2018}]{Wang2018ESRGANES}
\begin{bchapter}
\bauthor{\bsnm{Wang}, \binits{X.}},
\bauthor{\bsnm{Yu}, \binits{K.}},
\bauthor{\bsnm{Wu}, \binits{S.}},
\bauthor{\bsnm{Gu}, \binits{J.}},
\bauthor{\bsnm{Liu}, \binits{Y.}},
\bauthor{\bsnm{Dong}, \binits{C.}},
\bauthor{\bsnm{Loy}, \binits{C.C.}},
\bauthor{\bsnm{Qiao}, \binits{Y.}},
\bauthor{\bsnm{Tang}, \binits{X.}}:
\bctitle{Esrgan: Enhanced super-resolution generative adversarial networks}.
In: \bbtitle{ECCV Workshops}
(\byear{2018})
\end{bchapter}
\endbibitem

\bibitem[\protect\citeauthoryear{Ronneberger et~al.}{2015}]{RFB15a}
\begin{bchapter}
\bauthor{\bsnm{Ronneberger}, \binits{O.}},
\bauthor{\bsnm{P.Fischer}},
\bauthor{\bsnm{Brox}, \binits{T.}}:
\bctitle{U-net: Convolutional networks for biomedical image segmentation}.
In: \bbtitle{Medical Image Computing and Computer-Assisted Intervention
  (MICCAI)}
(\byear{2015})
\end{bchapter}
\endbibitem

\bibitem[\protect\citeauthoryear{Kim et~al.}{2022}]{Mssnet}
\begin{botherref}
\oauthor{\bsnm{Kim}, \binits{K.}},
\oauthor{\bsnm{Lee}, \binits{S.}},
\oauthor{\bsnm{Cho}, \binits{S.}}:
Mssnet: Multi-scale-stage network for single image deblurring.
ArXiv
\textbf{abs/2202.09652}
(2022)
\end{botherref}
\endbibitem

\bibitem[\protect\citeauthoryear{Dong et~al.}{2020}]{9156921}
\begin{bchapter}
\bauthor{\bsnm{Dong}, \binits{H.}},
\bauthor{\bsnm{Pan}, \binits{J.}},
\bauthor{\bsnm{Xiang}, \binits{L.}},
\bauthor{\bsnm{Hu}, \binits{Z.}},
\bauthor{\bsnm{Zhang}, \binits{X.}},
\bauthor{\bsnm{Wang}, \binits{F.}},
\bauthor{\bsnm{Yang}, \binits{M.-H.}}:
\bctitle{Multi-scale boosted dehazing network with dense feature fusion}.
In: \bbtitle{2020 IEEE/CVF Conference on Computer Vision and Pattern
  Recognition (CVPR)},
pp. \bfpage{2154}--\blpage{2164}
(\byear{2020}).
\doiurl{10.1109/CVPR42600.2020.00223}
\end{bchapter}
\endbibitem

\bibitem[\protect\citeauthoryear{Kupyn et~al.}{2017}]{Degan}
\begin{botherref}
\oauthor{\bsnm{Kupyn}, \binits{O.}},
\oauthor{\bsnm{Budzan}, \binits{V.}},
\oauthor{\bsnm{Mykhailych}, \binits{M.}},
\oauthor{\bsnm{Mishkin}, \binits{D.}},
\oauthor{\bsnm{Matas}, \binits{J.}}:
Deblurgan: Blind motion deblurring using conditional adversarial networks.
2018 IEEE/CVF Conference on Computer Vision and Pattern Recognition,
8183--8192
(2017)
\end{botherref}
\endbibitem

\bibitem[\protect\citeauthoryear{Zhang et~al.}{2020}]{DBGAN}
\begin{botherref}
\oauthor{\bsnm{Zhang}, \binits{K.}},
\oauthor{\bsnm{Luo}, \binits{W.}},
\oauthor{\bsnm{Zhong}, \binits{Y.}},
\oauthor{\bsnm{Ma}, \binits{L.}},
\oauthor{\bsnm{Stenger}, \binits{B.}},
\oauthor{\bsnm{Liu}, \binits{W.}},
\oauthor{\bsnm{Li}, \binits{H.}}:
Deblurring by realistic blurring.
2020 IEEE/CVF Conference on Computer Vision and Pattern Recognition (CVPR),
2734--2743
(2020)
\end{botherref}
\endbibitem

\bibitem[\protect\citeauthoryear{Kupyn et~al.}{2019}]{deganv2}
\begin{botherref}
\oauthor{\bsnm{Kupyn}, \binits{O.}},
\oauthor{\bsnm{Martyniuk}, \binits{T.}},
\oauthor{\bsnm{Wu}, \binits{J.}},
\oauthor{\bsnm{Wang}, \binits{Z.}}:
Deblurgan-v2: Deblurring (orders-of-magnitude) faster and better.
2019 IEEE/CVF International Conference on Computer Vision (ICCV),
8877--8886
(2019)
\end{botherref}
\endbibitem

\bibitem[\protect\citeauthoryear{Anwar and Barnes}{2020}]{r7anwar2019drln}
\begin{botherref}
\oauthor{\bsnm{Anwar}, \binits{S.}},
\oauthor{\bsnm{Barnes}, \binits{N.}}:
Densely residual laplacian super-resolution.
TPAMI
(2020)
\end{botherref}
\endbibitem

\bibitem[\protect\citeauthoryear{Zhang et~al.}{2018}]{r105zhang2018rcan}
\begin{bchapter}
\bauthor{\bsnm{Zhang}, \binits{Y.}},
\bauthor{\bsnm{Li}, \binits{K.}},
\bauthor{\bsnm{Li}, \binits{K.}},
\bauthor{\bsnm{Wang}, \binits{L.}},
\bauthor{\bsnm{Zhong}, \binits{B.}},
\bauthor{\bsnm{Fu}, \binits{Y.}}:
\bctitle{Image super-resolution using very deep residual channel attention
  networks}.
In: \bbtitle{ECCV}
(\byear{2018})
\end{bchapter}
\endbibitem

\bibitem[\protect\citeauthoryear{Zhang et~al.}{2020}]{r107zhang2020rdnir}
\begin{botherref}
\oauthor{\bsnm{Zhang}, \binits{Y.}},
\oauthor{\bsnm{Tian}, \binits{Y.}},
\oauthor{\bsnm{Kong}, \binits{Y.}},
\oauthor{\bsnm{Zhong}, \binits{B.}},
\oauthor{\bsnm{Fu}, \binits{Y.}}:
Residual dense network for image restoration.
TPAMI
(2020)
\end{botherref}
\endbibitem

\bibitem[\protect\citeauthoryear{Dudhane et~al.}{2022}]{r18dudhane2021burst}
\begin{bchapter}
\bauthor{\bsnm{Dudhane}, \binits{A.}},
\bauthor{\bsnm{Zamir}, \binits{S.W.}},
\bauthor{\bsnm{Khan}, \binits{S.}},
\bauthor{\bsnm{Khan}, \binits{F.S.}},
\bauthor{\bsnm{Yang}, \binits{M.-H.}}:
\bctitle{Burst image restoration and enhancement}.
In: \bbtitle{CVPR}
(\byear{2022})
\end{bchapter}
\endbibitem

\bibitem[\protect\citeauthoryear{Zamir et~al.}{2022}]{Zamir2022MIRNetv2}
\begin{botherref}
\oauthor{\bsnm{Zamir}, \binits{S.W.}},
\oauthor{\bsnm{Arora}, \binits{A.}},
\oauthor{\bsnm{Khan}, \binits{S.}},
\oauthor{\bsnm{Hayat}, \binits{M.}},
\oauthor{\bsnm{Khan}, \binits{F.S.}},
\oauthor{\bsnm{Yang}, \binits{M.-H.}},
\oauthor{\bsnm{Shao}, \binits{L.}}:
Learning enriched features for fast image restoration and enhancement.
TPAMI
(2022)
\end{botherref}
\endbibitem

\bibitem[\protect\citeauthoryear{Conde et~al.}{2022}]{conde2022swin2sr}
\begin{bchapter}
\bauthor{\bsnm{Conde}, \binits{M.V.}},
\bauthor{\bsnm{Choi}, \binits{U.-J.}},
\bauthor{\bsnm{Burchi}, \binits{M.}},
\bauthor{\bsnm{Timofte}, \binits{R.}}:
\bctitle{{S}win2{SR}: Swinv2 transformer for compressed image super-resolution
  and restoration}.
In: \bbtitle{Proceedings of the European Conference on Computer Vision (ECCV)
  Workshops}
(\byear{2022})
\end{bchapter}
\endbibitem

\bibitem[\protect\citeauthoryear{Liang et~al.}{2021}]{liang2021swinir}
\begin{botherref}
\oauthor{\bsnm{Liang}, \binits{J.}},
\oauthor{\bsnm{Cao}, \binits{J.}},
\oauthor{\bsnm{Sun}, \binits{G.}},
\oauthor{\bsnm{Zhang}, \binits{K.}},
\oauthor{\bsnm{Van~Gool}, \binits{L.}},
\oauthor{\bsnm{Timofte}, \binits{R.}}:
Swinir: Image restoration using swin transformer.
arXiv preprint arXiv:2108.10257
(2021)
\end{botherref}
\endbibitem

\bibitem[\protect\citeauthoryear{Cho et~al.}{2021}]{2021Rethinking}
\begin{bchapter}
\bauthor{\bsnm{Cho}, \binits{S.J.}},
\bauthor{\bsnm{Ji}, \binits{S.W.}},
\bauthor{\bsnm{Hong}, \binits{J.P.}},
\bauthor{\bsnm{Jung}, \binits{S.W.}},
\bauthor{\bsnm{Ko}, \binits{S.J.}}:
\bctitle{Rethinking coarse-to-fine approach in single image deblurring}.
In: \bbtitle{ICCV}
(\byear{2021})
\end{bchapter}
\endbibitem

\bibitem[\protect\citeauthoryear{Yue et~al.}{2020}]{r90ECCV2020_984}
\begin{bchapter}
\bauthor{\bsnm{Yue}, \binits{Z.}},
\bauthor{\bsnm{Zhao}, \binits{Q.}},
\bauthor{\bsnm{Zhang}, \binits{L.}},
\bauthor{\bsnm{Meng}, \binits{D.}}:
\bctitle{Dual adversarial network: Toward real-world noise removal and noise
  generation}.
In: \bbtitle{ECCV},
(\byear{2020})
\end{bchapter}
\endbibitem

\bibitem[\protect\citeauthoryear{Zhang
  et~al.}{2020}]{r99Zhang2020PlugandPlayIR}
\begin{barticle}
\bauthor{\bsnm{Zhang}, \binits{K.}},
\bauthor{\bsnm{Li}, \binits{Y.}},
\bauthor{\bsnm{Zuo}, \binits{W.}},
\bauthor{\bsnm{Zhang}, \binits{L.}},
\bauthor{\bsnm{Gool}, \binits{L.V.}},
\bauthor{\bsnm{Timofte}, \binits{R.}}:
\batitle{Plug-and-play image restoration with deep denoiser prior}.
\bjtitle{TPAMI}
\bvolume{44},
\bfpage{6360}--\blpage{6376}
(\byear{2020})
\end{barticle}
\endbibitem

\bibitem[\protect\citeauthoryear{Zhu et~al.}{2018}]{2018DehazeGAN}
\begin{bchapter}
\bauthor{\bsnm{Zhu}, \binits{H.}},
\bauthor{\bsnm{Xi}, \binits{P.}},
\bauthor{\bsnm{Chandrasekhar}, \binits{V.}},
\bauthor{\bsnm{Li}, \binits{L.}},
\bauthor{\bsnm{Lim}, \binits{J.H.}}:
\bctitle{Dehazegan: When image dehazing meets differential programming}.
In: \bbtitle{Twenty-Seventh International Joint Conference on Artificial
  Intelligence IJCAI-18}
(\byear{2018})
\end{bchapter}
\endbibitem

\bibitem[\protect\citeauthoryear{Guo et~al.}{2019}]{2019Dense}
\begin{bchapter}
\bauthor{\bsnm{Guo}, \binits{T.}},
\bauthor{\bsnm{Li}, \binits{X.}},
\bauthor{\bsnm{Cherukuri}, \binits{V.}},
\bauthor{\bsnm{Monga}, \binits{V.}}:
\bctitle{Dense scene information estimation network for dehazing}.
In: \bbtitle{2019 IEEE/CVF Conference on Computer Vision and Pattern
  Recognition Workshops (CVPRW)}
(\byear{2019})
\end{bchapter}
\endbibitem

\bibitem[\protect\citeauthoryear{Yang et~al.}{2019}]{2019Dual}
\begin{bchapter}
\bauthor{\bsnm{Yang}, \binits{A.}},
\bauthor{\bsnm{Wang}, \binits{H.}},
\bauthor{\bsnm{Ji}, \binits{Z.}},
\bauthor{\bsnm{Pang}, \binits{Y.}},
\bauthor{\bsnm{Shao}, \binits{L.}}:
\bctitle{Dual-path in dual-path network for single image dehazing}.
In: \bbtitle{Twenty-Eighth International Joint Conference on Artificial
  Intelligence {IJCAI-19}}
(\byear{2019})
\end{bchapter}
\endbibitem

\bibitem[\protect\citeauthoryear{Tian et~al.}{2021}]{tian2021designing}
\begin{barticle}
\bauthor{\bsnm{Tian}, \binits{C.}},
\bauthor{\bsnm{Xu}, \binits{Y.}},
\bauthor{\bsnm{Zuo}, \binits{W.}},
\bauthor{\bsnm{Du}, \binits{B.}},
\bauthor{\bsnm{Lin}, \binits{C.-W.}},
\bauthor{\bsnm{Zhang}, \binits{D.}}:
\batitle{Designing and training of a dual cnn for image denoising}.
\bjtitle{Knowledge-Based Systems}
\bvolume{226},
\bfpage{106949}
(\byear{2021})
\end{barticle}
\endbibitem

\bibitem[\protect\citeauthoryear{Singh et~al.}{2020}]{2020Refining}
\begin{botherref}
\oauthor{\bsnm{Singh}, \binits{V.}},
\oauthor{\bsnm{Ramnath}, \binits{K.}},
\oauthor{\bsnm{Mittal}, \binits{A.}}:
Refining high-frequencies for sharper super-resolution and deblurring.
Computer Vision and Image Understanding
(2020)
\end{botherref}
\endbibitem

\bibitem[\protect\citeauthoryear{Pan et~al.}{2018}]{2018LearningD}
\begin{bchapter}
\bauthor{\bsnm{Pan}, \binits{J.}},
\bauthor{\bsnm{Liu}, \binits{S.}},
\bauthor{\bsnm{Sun}, \binits{D.}},
\bauthor{\bsnm{Zhang}, \binits{J.}},
\bauthor{\bsnm{Liu}, \binits{Y.}},
\bauthor{\bsnm{Ren}, \binits{J.}},
\bauthor{\bsnm{Li}, \binits{Z.}},
\bauthor{\bsnm{Tang}, \binits{J.}},
\bauthor{\bsnm{Lu}, \binits{H.}},
\bauthor{\bsnm{Tai}, \binits{Y.W.a.}}:
\bctitle{Learning dual convolutional neural networks for low-level vision}.
In: \bbtitle{CVPR}
(\byear{2018})
\end{bchapter}
\endbibitem

\bibitem[\protect\citeauthoryear{Siyuan et~al.}{2018}]{2018Fast}
\begin{botherref}
\oauthor{\bsnm{Siyuan}, \binits{L.I.}},
\oauthor{\bsnm{Ren}, \binits{W.}},
\oauthor{\bsnm{Zhang}, \binits{J.}},
\oauthor{\bsnm{Yu}, \binits{J.}},
\oauthor{\bsnm{Guo}, \binits{X.}}:
Fast Single Image Rain Removal via a Deep Decomposition-Composition Network.
Computer Vision and Image Understanding
(2018)
\end{botherref}
\endbibitem

\bibitem[\protect\citeauthoryear{Li et~al.}{2019}]{Li2019RethinkingOM}
\begin{botherref}
\oauthor{\bsnm{Li}, \binits{W.}},
\oauthor{\bsnm{Wang}, \binits{Z.}},
\oauthor{\bsnm{Yin}, \binits{B.}},
\oauthor{\bsnm{Peng}, \binits{Q.}},
\oauthor{\bsnm{Du}, \binits{Y.}},
\oauthor{\bsnm{Xiao}, \binits{T.}},
\oauthor{\bsnm{Yu}, \binits{G.}},
\oauthor{\bsnm{Lu}, \binits{H.}},
\oauthor{\bsnm{Wei}, \binits{Y.}},
\oauthor{\bsnm{Sun}, \binits{J.}}:
Rethinking on multi-stage networks for human pose estimation.
ArXiv
\textbf{abs/1901.00148}
(2019)
\end{botherref}
\endbibitem

\bibitem[\protect\citeauthoryear{Cheng et~al.}{2019}]{15Cheng2019SPGNetSP}
\begin{botherref}
\oauthor{\bsnm{Cheng}, \binits{B.}},
\oauthor{\bsnm{Chen}, \binits{L.-C.}},
\oauthor{\bsnm{Wei}, \binits{Y.}},
\oauthor{\bsnm{Zhu}, \binits{Y.}},
\oauthor{\bsnm{Huang}, \binits{Z.}},
\oauthor{\bsnm{Xiong}, \binits{J.}},
\oauthor{\bsnm{Huang}, \binits{T.}},
\oauthor{\bsnm{Hwu}, \binits{W.-m.W.}},
\oauthor{\bsnm{Shi}, \binits{H.}}:
Spgnet: Semantic prediction guidance for scene parsing.
2019 IEEE/CVF International Conference on Computer Vision (ICCV),
5217--5227
(2019)
\end{botherref}
\endbibitem

\bibitem[\protect\citeauthoryear{Ghosh et~al.}{2018}]{26Ghosh2018StackedSG}
\begin{botherref}
\oauthor{\bsnm{Ghosh}, \binits{P.}},
\oauthor{\bsnm{Yao}, \binits{Y.}},
\oauthor{\bsnm{Davis}, \binits{L.S.}},
\oauthor{\bsnm{Divakaran}, \binits{A.}}:
Stacked spatio-temporal graph convolutional networks for action segmentation.
2020 IEEE Winter Conference on Applications of Computer Vision (WACV),
565--574
(2018)
\end{botherref}
\endbibitem

\bibitem[\protect\citeauthoryear{Li et~al.}{2020}]{45li2020ms}
\begin{botherref}
\oauthor{\bsnm{Li}, \binits{S.-J.}},
\oauthor{\bsnm{AbuFarha}, \binits{Y.}},
\oauthor{\bsnm{Liu}, \binits{Y.}},
\oauthor{\bsnm{Cheng}, \binits{M.-M.}},
\oauthor{\bsnm{Gall}, \binits{J.}}:
Ms-tcn++: Multi-stage temporal convolutional network for action segmentation.
IEEE Transactions on Pattern Analysis and Machine Intelligence,
1--1
(2020)
\doiurl{10.1109/TPAMI.2020.3021756}
\end{botherref}
\endbibitem

\bibitem[\protect\citeauthoryear{Tao et~al.}{2018}]{2018Scale}
\begin{botherref}
\oauthor{\bsnm{Tao}, \binits{X.}},
\oauthor{\bsnm{Gao}, \binits{H.}},
\oauthor{\bsnm{Wang}, \binits{Y.}},
\oauthor{\bsnm{Shen}, \binits{X.}},
\oauthor{\bsnm{Wang}, \binits{J.}},
\oauthor{\bsnm{Jia}, \binits{J.}}:
Scale-recurrent network for deep image deblurring.
CVPR
(2018)
\end{botherref}
\endbibitem

\bibitem[\protect\citeauthoryear{Fu et~al.}{2018}]{2018Lightweight}
\begin{botherref}
\oauthor{\bsnm{Fu}, \binits{X.}},
\oauthor{\bsnm{Liang}, \binits{B.}},
\oauthor{\bsnm{Huang}, \binits{Y.}},
\oauthor{\bsnm{Ding}, \binits{X.}},
\oauthor{\bsnm{Paisley}, \binits{J.}}:
Lightweight pyramid networks for image deraining.
IEEE Transactions on Neural Networks and Learning Systems
(2018)
\end{botherref}
\endbibitem

\bibitem[\protect\citeauthoryear{Zhang et~al.}{2022}]{zhang2022event}
\begin{botherref}
\oauthor{\bsnm{Zhang}, \binits{H.}},
\oauthor{\bsnm{Zhang}, \binits{L.}},
\oauthor{\bsnm{Dai}, \binits{Y.}},
\oauthor{\bsnm{Li}, \binits{H.}},
\oauthor{\bsnm{Koniusz}, \binits{P.}}:
Event-guided multi-patch network with self-supervision for non-uniform motion
  deblurring.
International Journal of Computer Vision,
1--18
(2022)
\end{botherref}
\endbibitem

\bibitem[\protect\citeauthoryear{Zhang et~al.}{2019}]{Zhang_2019_CVPR}
\begin{bchapter}
\bauthor{\bsnm{Zhang}, \binits{H.}},
\bauthor{\bsnm{Dai}, \binits{Y.}},
\bauthor{\bsnm{Li}, \binits{H.}},
\bauthor{\bsnm{Koniusz}, \binits{P.}}:
\bctitle{Deep stacked hierarchical multi-patch network for image deblurring}.
In: \bbtitle{The IEEE Conference on Computer Vision and Pattern Recognition
  (CVPR)}
(\byear{2019})
\end{bchapter}
\endbibitem

\bibitem[\protect\citeauthoryear{Fu et~al.}{2017}]{238099669}
\begin{bchapter}
\bauthor{\bsnm{Fu}, \binits{X.}},
\bauthor{\bsnm{Huang}, \binits{J.}},
\bauthor{\bsnm{Zeng}, \binits{D.}},
\bauthor{\bsnm{Huang}, \binits{Y.}},
\bauthor{\bsnm{Ding}, \binits{X.}},
\bauthor{\bsnm{Paisley}, \binits{J.}}:
\bctitle{Removing rain from single images via a deep detail network}.
In: \bbtitle{CVPR}
(\byear{2017})
\end{bchapter}
\endbibitem

\bibitem[\protect\citeauthoryear{Yang et~al.}{2017}]{81Yang2016DeepJR}
\begin{botherref}
\oauthor{\bsnm{Yang}, \binits{W.}},
\oauthor{\bsnm{Tan}, \binits{R.T.}},
\oauthor{\bsnm{Feng}, \binits{J.}},
\oauthor{\bsnm{Liu}, \binits{J.}},
\oauthor{\bsnm{Guo}, \binits{Z.}},
\oauthor{\bsnm{Yan}, \binits{S.}}:
Deep joint rain detection and removal from a single image.
CVPR
(2017)
\end{botherref}
\endbibitem

\bibitem[\protect\citeauthoryear{Zhang et~al.}{2017}]{90Zhang2017ImageDU}
\begin{barticle}
\bauthor{\bsnm{Zhang}, \binits{H.}},
\bauthor{\bsnm{Sindagi}, \binits{V.A.}},
\bauthor{\bsnm{Patel}, \binits{V.M.}}:
\batitle{Image de-raining using a conditional generative adversarial network}.
\bjtitle{IEEE Transactions on Circuits and Systems for Video Technology}
\bvolume{30},
\bfpage{3943}--\blpage{3956}
(\byear{2017})
\end{barticle}
\endbibitem

\bibitem[\protect\citeauthoryear{Zhang and Patel}{2018}]{DIDMDN}
\begin{botherref}
\oauthor{\bsnm{Zhang}, \binits{H.}},
\oauthor{\bsnm{Patel}, \binits{V.M.}}:
Density-aware single image de-raining using a multi-stream dense network.
2018 IEEE/CVF Conference on Computer Vision and Pattern Recognition,
695--704
(2018)
\end{botherref}
\endbibitem

\bibitem[\protect\citeauthoryear{Nah et~al.}{2016}]{Gopro}
\begin{botherref}
\oauthor{\bsnm{Nah}, \binits{S.}},
\oauthor{\bsnm{Kim}, \binits{T.H.}},
\oauthor{\bsnm{Lee}, \binits{K.M.}}:
Deep multi-scale convolutional neural network for dynamic scene deblurring.
2017 IEEE Conference on Computer Vision and Pattern Recognition (CVPR),
257--265
(2016)
\end{botherref}
\endbibitem

\bibitem[\protect\citeauthoryear{Li et~al.}{2016}]{487780668}
\begin{bchapter}
\bauthor{\bsnm{Li}, \binits{Y.}},
\bauthor{\bsnm{Tan}, \binits{R.T.}},
\bauthor{\bsnm{Guo}, \binits{X.}},
\bauthor{\bsnm{Lu}, \binits{J.}},
\bauthor{\bsnm{Brown}, \binits{M.S.}}:
\bctitle{Rain streak removal using layer priors}.
In: \bbtitle{CVPR}
(\byear{2016})
\end{bchapter}
\endbibitem

\bibitem[\protect\citeauthoryear{Shen et~al.}{2019}]{HIDE}
\begin{botherref}
\oauthor{\bsnm{Shen}, \binits{Z.}},
\oauthor{\bsnm{Wang}, \binits{W.}},
\oauthor{\bsnm{Lu}, \binits{X.}},
\oauthor{\bsnm{Shen}, \binits{J.}},
\oauthor{\bsnm{Ling}, \binits{H.}},
\oauthor{\bsnm{Xu}, \binits{T.}},
\oauthor{\bsnm{Shao}, \binits{L.}}:
Human-aware motion deblurring.
2019 IEEE/CVF International Conference on Computer Vision (ICCV),
5571--5580
(2019)
\end{botherref}
\endbibitem

\bibitem[\protect\citeauthoryear{Kingma and Ba}{2014}]{2014Adam}
\begin{botherref}
\oauthor{\bsnm{Kingma}, \binits{D.}},
\oauthor{\bsnm{Ba}, \binits{J.}}:
Adam: A method for stochastic optimization.
Computer Science
(2014)
\end{botherref}
\endbibitem

\bibitem[\protect\citeauthoryear{Loshchilov and Hutter}{2016}]{2016SGDR}
\begin{botherref}
\oauthor{\bsnm{Loshchilov}, \binits{I.}},
\oauthor{\bsnm{Hutter}, \binits{F.}}:
Sgdr: Stochastic gradient descent with warm restarts
(2016)
\end{botherref}
\endbibitem

\bibitem[\protect\citeauthoryear{Chu et~al.}{2021}]{Chu2021ImprovingIR}
\begin{bchapter}
\bauthor{\bsnm{Chu}, \binits{X.}},
\bauthor{\bsnm{Chen}, \binits{L.}},
\bauthor{\bsnm{Chen}, \binits{C.}},
\bauthor{\bsnm{Lu}, \binits{X.}}:
\bctitle{Improving image restoration by revisiting global information
  aggregation}.
In: \bbtitle{ECCV}
(\byear{2021})
\end{bchapter}
\endbibitem

\bibitem[\protect\citeauthoryear{Jiang et~al.}{2020}]{MSPFN}
\begin{botherref}
\oauthor{\bsnm{Jiang}, \binits{K.}},
\oauthor{\bsnm{Wang}, \binits{Z.}},
\oauthor{\bsnm{Yi}, \binits{P.}},
\oauthor{\bsnm{Chen}, \binits{C.}},
\oauthor{\bsnm{Huang}, \binits{B.}},
\oauthor{\bsnm{Luo}, \binits{Y.}},
\oauthor{\bsnm{Ma}, \binits{J.}},
\oauthor{\bsnm{Jiang}, \binits{J.}}:
Multi-scale progressive fusion network for single image deraining.
CVPR
(2020)
\end{botherref}
\endbibitem

\bibitem[\protect\citeauthoryear{Purohit et~al.}{2021}]{SPAIR}
\begin{botherref}
\oauthor{\bsnm{Purohit}, \binits{K.}},
\oauthor{\bsnm{Suin}, \binits{M.}},
\oauthor{\bsnm{Rajagopalan}, \binits{A.N.}},
\oauthor{\bsnm{Boddeti}, \binits{V.N.}}:
Spatially-adaptive image restoration using distortion-guided networks.
CoRR
\textbf{abs/2108.08617}
(2021)
\end{botherref}
\endbibitem

\bibitem[\protect\citeauthoryear{Zhang et~al.}{2017}]{Test100}
\begin{barticle}
\bauthor{\bsnm{Zhang}, \binits{H.}},
\bauthor{\bsnm{Sindagi}, \binits{V.A.}},
\bauthor{\bsnm{Patel}, \binits{V.M.}}:
\batitle{Image de-raining using a conditional generative adversarial network}.
\bjtitle{IEEE Transactions on Circuits and Systems for Video Technology}
\bvolume{30},
\bfpage{3943}--\blpage{3956}
(\byear{2017})
\end{barticle}
\endbibitem

\bibitem[\protect\citeauthoryear{Yang et~al.}{2016}]{Rain100}
\begin{botherref}
\oauthor{\bsnm{Yang}, \binits{W.}},
\oauthor{\bsnm{Tan}, \binits{R.T.}},
\oauthor{\bsnm{Feng}, \binits{J.}},
\oauthor{\bsnm{Liu}, \binits{J.}},
\oauthor{\bsnm{Guo}, \binits{Z.}},
\oauthor{\bsnm{Yan}, \binits{S.}}:
Deep joint rain detection and removal from a single image.
2017 IEEE Conference on Computer Vision and Pattern Recognition (CVPR),
1685--1694
(2016)
\end{botherref}
\endbibitem

\bibitem[\protect\citeauthoryear{Fu et~al.}{2016}]{DerainNet}
\begin{barticle}
\bauthor{\bsnm{Fu}, \binits{X.}},
\bauthor{\bsnm{Huang}, \binits{J.}},
\bauthor{\bsnm{Ding}, \binits{X.}},
\bauthor{\bsnm{Liao}, \binits{Y.}},
\bauthor{\bsnm{Paisley}, \binits{J.W.}}:
\batitle{Clearing the skies: A deep network architecture for single-image rain
  removal}.
\bjtitle{TIP}
\bvolume{26},
\bfpage{2944}--\blpage{2956}
(\byear{2016})
\end{barticle}
\endbibitem

\bibitem[\protect\citeauthoryear{Wei et~al.}{2018}]{semi}
\begin{botherref}
\oauthor{\bsnm{Wei}, \binits{W.}},
\oauthor{\bsnm{Meng}, \binits{D.}},
\oauthor{\bsnm{Zhao}, \binits{Q.}},
\oauthor{\bsnm{Xu}, \binits{Z.}}:
Semi-supervised cnn for single image rain removal.
ArXiv
\textbf{abs/1807.11078}
(2018)
\end{botherref}
\endbibitem

\bibitem[\protect\citeauthoryear{Yasarla and Patel}{2019}]{UMRL}
\begin{botherref}
\oauthor{\bsnm{Yasarla}, \binits{R.}},
\oauthor{\bsnm{Patel}, \binits{V.M.}}:
Uncertainty guided multi-scale residual learning-using a cycle spinning cnn for
  single image de-raining.
2019 IEEE/CVF Conference on Computer Vision and Pattern Recognition (CVPR),
8397--8406
(2019)
\end{botherref}
\endbibitem

\bibitem[\protect\citeauthoryear{Gao et~al.}{2019}]{Gao2019DynamicSD}
\begin{botherref}
\oauthor{\bsnm{Gao}, \binits{H.}},
\oauthor{\bsnm{Tao}, \binits{X.}},
\oauthor{\bsnm{Shen}, \binits{X.}},
\oauthor{\bsnm{Jia}, \binits{J.}}:
Dynamic scene deblurring with parameter selective sharing and nested skip
  connections.
2019 IEEE/CVF Conference on Computer Vision and Pattern Recognition (CVPR),
3843--3851
(2019)
\end{botherref}
\endbibitem

\bibitem[\protect\citeauthoryear{Park et~al.}{2019}]{MTRNN}
\begin{bchapter}
\bauthor{\bsnm{Park}, \binits{D.}},
\bauthor{\bsnm{Kang}, \binits{D.U.}},
\bauthor{\bsnm{Kim}, \binits{J.}},
\bauthor{\bsnm{Chun}, \binits{S.Y.}}:
\bctitle{Multi-temporal recurrent neural networks for progressive non-uniform
  single image deblurring with incremental temporal training}.
In: \bbtitle{ECCV}
(\byear{2019})
\end{bchapter}
\endbibitem

\bibitem[\protect\citeauthoryear{Suin
  et~al.}{2020}]{Suin2020SpatiallyAttentivePN}
\begin{botherref}
\oauthor{\bsnm{Suin}, \binits{M.}},
\oauthor{\bsnm{Purohit}, \binits{K.}},
\oauthor{\bsnm{Rajagopalan}, \binits{A.N.}}:
Spatially-attentive patch-hierarchical network for adaptive motion deblurring.
CVPR,
3603--3612
(2020)
\end{botherref}
\endbibitem

\end{thebibliography}

\end{document}